\documentclass[twoside,11pt]{article}

\usepackage{jmlr2e}
\usepackage{mathrsfs,amsmath,amsfonts,amssymb,bm,bbm,eufrak,dsfont,pifont,epsfig,euscript,stmaryrd}

\newcommand{\bb}{\mathbb}
\newcommand{\eu}{\EuScript}

\ShortHeadings{Universality, Characteristic Kernels and RKHS Embedding of Measures}{Sriperumbudur, Fukumizu and Lanckriet}
\firstpageno{1}

\begin{document}

\title{Universality, Characteristic Kernels and RKHS Embedding of Measures}

\author{\name Bharath K. Sriperumbudur \email bharathsv@ucsd.edu \\
       \addr Department of Electrical and Computer Engineering\\
       University of California, San Diego\\
       La Jolla, CA 92093-0407, USA.
       \AND
		 \name Kenji Fukumizu \email fukumizu@ism.ac.jp\\
		 \addr The Institute of Statistical Mathematics\\
		 10-3 Midori-cho, Tachikawa\\
		 Tokyo 190-8562, Japan.
		 \AND
		 \name Gert R. G. Lanckriet \email gert@ece.ucsd.edu \\
       \addr Department of Electrical and Computer Engineering\\
       University of California, San Diego\\
       La Jolla, CA 92093-0407, USA.
}

\editor{}

\maketitle

\begin{abstract}
A Hilbert space embedding for probability measures has recently been proposed, wherein any probability measure is represented as a mean element in a reproducing kernel Hilbert space (RKHS). Such an embedding has found applications in homogeneity testing, independence testing, dimensionality reduction, etc., with the requirement that the reproducing kernel is \emph{characteristic}, i.e., the embedding is injective.
\par In this paper, we generalize this embedding to finite signed Borel measures, wherein any finite signed Borel measure is represented as a mean element in an RKHS. We show that the proposed embedding is injective if and only if the kernel is \emph{universal}. This therefore, provides a novel characterization of universal kernels, which are proposed in the context of achieving the \emph{Bayes risk} by kernel-based classification/regression algorithms. By exploiting this relation between universality and the embedding of finite signed Borel measures into an RKHS, we establish the relation between universal and characteristic kernels.
\end{abstract}


\section{Introduction}\label{Sec:Introduction}
Kernel methods have been popular in machine learning and pattern analysis for their superior performance on a wide spectrum of learning tasks. They are broadly established as an easy way to construct nonlinear algorithms from linear ones, by embedding data points into higher dimensional reproducing kernel Hilbert spaces (RKHSs) \citep{Scholkopf-02,ShaweTaylor-04}. Recently, this idea has been generalized to embed probability distributions into RKHSs, which provides a linear method for dealing with higher order statistics \citep{Gretton-06,Smola-07,Fukumizu-08a,Fukumizu-08b,Sriperumbudur-08,Sriperumbudur-09c,Sriperumbudur-09a}. Formally, given the set of all Borel probability measures defined on the topological space $X$, and the RKHS $(\eu{H},k)$ of functions on $X$ with $k:X\times X\rightarrow\bb{R}$ as its reproducing kernel (r.k.) that is measurable and bounded, any Borel probability measure, $\bb{P}$ is embedded as,
\begin{equation}\label{Eq:prob}
\bb{P}\mapsto\int_X k(\cdot,x)\,d\bb{P}(x).
\end{equation}
Such an embedding has been found to be useful in many statistical applications like homogeneity testing \citep{Gretton-06}, independence testing \citep{Gretton-08,Fukumizu-08a}, dimensionality reduction \citep{Fukumizu-04,Fukumizu-09}, etc., as it provides a powerful and straightforward method of dealing with higher-order statistics of random variables. However, in these applications, it is critical that the embedding in (\ref{Eq:prob}) is injective so that probability measures can be distinguished by their images in $\eu{H}$. To this end, \citet{Fukumizu-08a} introduced the notion of \emph{characteristic kernel}~---~a bounded, measurable $k$ is said to be \emph{characteristic} if (\ref{Eq:prob}) is injective~---~for which many characterizations have recently been provided \citep{Gretton-06,Fukumizu-08a,Fukumizu-08b,Sriperumbudur-08,Sriperumbudur-09c,Sriperumbudur-09a}.
\par A natural extension to the above idea of embedding probability measures into an RKHS, $\eu{H}$ is to embed finite signed Borel measures, $\mu$ into $\eu{H}$ as 
\begin{equation}\label{Eq:signedmeasure}
\mu\mapsto\int_X k(\cdot,x)\,d\mu(x),
\end{equation}
and study the conditions on the kernel, $k$ for which such an embedding is injective. Although the embedding in (\ref{Eq:signedmeasure}) can be proposed and investigated for mathematical pleasure, we show as one of the main contributions of this paper that under certain conditions on $\mu$ and $X$, the embedding in (\ref{Eq:signedmeasure}) is closely related to the concept of \emph{universal kernels} (see Section~\ref{subsec:universal} for the formal introduction to universal kernels), which was first proposed by \citet{Steinwart-01}~---~in the context of achieving the \emph{Bayes risk} in kernel-based classification/regression algorithms~---~and later extended by \citet{Micchelli-06}, \citet{Carmeli-09} and  \citet{Sriperumbudur-09d}.\footnote{The present paper is an extended version of \citet{Sriperumbudur-09d}.} This connection shows that the embedding in (\ref{Eq:signedmeasure}) is not just an abstract mathematical object, but has applications in kernel-based classification/regression algorithms. Using the connection between (\ref{Eq:signedmeasure}) and universal kernels, we then show how the various notions of universality mentioned above are related to each other. In addition, since the embedding in (\ref{Eq:signedmeasure}) is a generalization of the embedding in (\ref{Eq:prob}), we also demonstrate the relation between characteristic kernels and universal kernels, which extends the preliminary study carried out in \citet[Section 3.4]{Sriperumbudur-09a}.

In the remainder of this introduction, we provide a comprehensive overview of our contributions which are presented in detail in later sections. First, in Section~\ref{subsec:universal}, we introduce universality, briefly discuss various notions of universality that are proposed in literature, and outline our contribution: a measure embedding view point of universality, which is novel and different from the existing view point of approximating functions in some target space by functions in an RKHS. We show that a kernel is universal if and only if the embedding in (\ref{Eq:signedmeasure}) is injective. Second, in Section~\ref{subsec:relation}, we discuss our second contribution of relating universal and characteristic kernels. 

\subsection{Contribution 1: Injective RKHS embedding of finite signed Radon measures to characterize universality}\label{subsec:universal}
In the regularization approach to learning \citep{Evgeniou-00}, it is well known that kernel-based algorithms (for classification/regression) generally invoke the \emph{representer theorem} \citep{Kimeldorf-70,Scholkopf-01} and learn a function in $\eu{H}$ that has the representation,
\begin{equation}\label{Eq:representer}
f:=\sum_{j\in\bb{N}_n} c_j k(\cdot,x_j),
\end{equation}
where $\bb{N}_n:=\{1,2,\ldots,n\}$ and $\{c_j:j\in\bb{N}_n\}\subset\bb{R}$ are parameters typically obtained from training data, $\{x_j:j\in\bb{N}_n\}\subset X$. As noted in \citet{Micchelli-06}, one can ask  whether the function, $f$ in (\ref{Eq:representer}) approximates any real-valued target function arbitrarily \emph{well} as the number of summands increases without bound. This is an important question to consider because if the answer is affirmative, then the kernel-based learning algorithm is \emph{consistent} in the sense that for any target function, $f^\star$ (which is usually assumed to belong to some subset of the space of real-valued continuous functions defined on $X$), the discrepancy between $f$ (which is learned from the training data) and $f^\star$ goes to zero (in some sense) as the sample size goes to infinity. Since 
\begin{equation}
\left\{\sum_{j\in\bb{N}_n}c_j k(\cdot,x_j):n\in\bb{N},\{c_j\}\subset\bb{R}, \{x_j\}\subset X\right\}\nonumber
\end{equation}
is dense in $\eu{H}$ \citep{Aronszajn-50}, and assuming that the kernel-based algorithm makes $f$ ``converge to an appropriate function" in $\eu{H}$ as $n\rightarrow \infty$, the above question of approximating $f^\star$ arbitrarily \emph{well} by $f$ in (\ref{Eq:representer}) as $n$ goes to infinity is equivalent to the question of \emph{whether $\eu{H}$ is rich enough to approximate any $f^\star$ arbitrarily well}, i.e., \emph{whether $\eu{H}$ is universal}. We show that characterizing universal RKHSs (or equivalently, the characterization of corresponding reproducing kernels (r.k.) as any RKHS is uniquely determined by its reproducing kernel) leads to the embedding in (\ref{Eq:signedmeasure}).
\par As mentioned above, the goal is to characterize $\eu{H}$ that allow to approximate any $f^\star$ in some target space, usually assumed to be some subset of the space of real-valued continuous functions on $X$. Therefore, depending on the choice of $X$, the choice of target space and the type of approximation, various notions of \emph{universality} have been proposed \citep{Steinwart-01,Micchelli-06,Carmeli-09,Sriperumbudur-09d}, which are briefly discussed in the following paragraphs. The eventual goal is to have a notion of universality that allows comprehensive (and general) necessary and/or sufficient conditions on the reproducing kernel for approximating, as strong as possible, a class of target functions, as general as possible.\vspace{2mm}
\par\noindent $c$-\textbf{universality:} Let $C(X)$ denote the space of continuous real-valued functions on some topological space, $X$. \citet{Steinwart-01} considered the above approximation problem when $X$ is a compact metric space, with $f^\star\in C(X)$ and defined a continuous kernel, $k$ as \emph{universal} (in this paper, we refer to it as \emph{c-universal}) if its associated RKHS, $\eu{H}$ is dense in $C(X)$ w.r.t.~the uniform norm (see Section~\ref{Sec:Notation} for the definition of uniform norm), i.e., for any $f^\star\in C(X)$, there exists a $g\in \eu{H}$ that uniformly approximates $f^\star$. In the context of learning, this indicates that if a kernel is \emph{c-universal}, then the corresponding kernel-based learning algorithm could be consistent in the sense that any target function, $f^\star\in C(X)$ could be approximated arbitrarily well in the uniform norm by $f$ in (\ref{Eq:representer}) as $n$ goes to infinity (see \citet[Corollary 5.29]{Steinwart-08} for a rigorous result). By applying the Stone-Weierstra\ss\;\,theorem \citep[Theorem 4.45]{Folland-99}, \citet{Steinwart-01} then provided sufficient conditions for a kernel to be \emph{c-universal}, using which the Gaussian kernel is shown to be \emph{c-universal} on every compact subset of $\bb{R}^d$. 
\par As our contribution, in Section~\ref{subsec:main}, we completely characterize \emph{c-universal} kernels by showing that $k$ is \emph{c-universal} if and only if the embedding in (\ref{Eq:signedmeasure}) is injective for $\mu\in M_b(X)$, the space of finite signed Radon measures defined on a compact Hausdorff space, $X$ (see Section~\ref{Sec:Notation} for 
a formal definition of $M_b(X)$). It has to be noted that this result is different from and more general~---~as both necessary and sufficient conditions are provided~---~than the one by \citet[Theorem 9]{Steinwart-01}, where only a sufficient condition is provided. Using this characterization, as a special case, we also obtain necessary and sufficient conditions for a \emph{Fourier kernel} (see Section~\ref{subsubsec:a2}) to be \emph{c-universal}, while \citet{Steinwart-01} provided only a sufficient condition.\vspace{2mm}
\par\noindent $cc$-\textbf{universality:} One limitation in the setup considered by \citet{Steinwart-01} is that $X$ is assumed to be compact, which excludes many interesting spaces, such as $\bb{R}^d$ and \emph{infinite} discrete sets. To overcome this limitation, \citet[Definition 2, Theorem 3]{Carmeli-09} and \citet{Sriperumbudur-09d} approximated any 
$f^\star\in C(X)$ by some $g\in\eu{H}$ uniformly over every compact set, $Z\subset X$, by defining a continuous kernel, $k$ to be \emph{universal} (in this paper, we refer to it as \emph{cc-universal}) if the corresponding RKHS, $\eu{H}$ is dense in $C(X)$ with the \emph{topology of compact convergence}, where $X$ is a non-compact Hausdorff space. I.e., for any compact set $Z\subset X$, for any $f^\star\in C(Z)$, there exists a $g\in \eu{H}_{\vert Z}$ that uniformly approximates $f^\star$. Here, $C(Z)$ is the space of all continuous real-valued functions on $Z$ equipped with the uniform norm, $\eu{H}_{\vert Z}:=\{f_{\vert Z}\,:\,f\in\eu{H}\}$ is the restriction of $\eu{H}$ to $Z$ and $f_{\vert Z}$ is the restriction of $f$ to $Z$.
\par As our contribution, in Section~\ref{subsec:main}, we show that $k$ is \emph{cc-universal} if and only if the embedding in (\ref{Eq:signedmeasure}) is injective for $\mu\in M_{bc}(X)$, the space of compactly supported finite signed Radon measures defined on a non-compact Hausdorff space, $X$. Compared to the characterization by \citet[Theorem 4]{Carmeli-09}, which deals with the injectivity of a certain integral operator on the space of square-integrable functions, our characterization is easy to understand~---~as it is related to a generalization of the 
embedding in (\ref{Eq:prob})~---~and will naturally lead to understanding the relation between \emph{cc-universal} and \emph{characteristic} kernels. Using this characterization, we also show that $k$ is \emph{cc-universal} if and only if it is universal in the sense of \citet{Micchelli-06}: for any compact $Z\subset X$, the set $K(Z):=\overline{\text{span}}\{k(\cdot,y):y\in Z\}$ is dense in $C(Z)$ in the uniform norm (see Remark~\ref{rem:universal}(b); also see \citet[Remark 1]{Carmeli-09}). As examples, many popular kernels on $\bb{R}^d$ are shown to be \emph{cc-universal} (see Sections~\ref{subsubsec:a1} and \ref{subsubsec:a3}; also see \citet[Section 4]{Micchelli-06}): Gaussian, Laplacian, $B_{2l+1}$-spline, sinc kernel, etc. \vspace{2mm}
\par\noindent $c_0$-\textbf{universality:} Although \emph{cc-universality} solves the limitation of \emph{c-universality} by handling non-compact $X$, the topology of compact convergence considered in \emph{cc-universality} is weaker than the topology of \emph{uniform convergence}, i.e., a sequence of functions, $\{f_n\}\subset C(X)$ converging to $f\in C(X)$ in the topology of uniform convergence ensures that they converge in the topology of compact convergence but not vice-versa. So, the natural question to ask is whether we can characterize $\eu{H}$ that are rich enough to approximate any $f^\star$ on non-compact $X$ in a stronger sense, i.e., uniformly, by some $g\in\eu{H}$. Recently, this has been answered by \citet[Definition 2, Theorem 1]{Carmeli-09} and \citet{Sriperumbudur-09d}, wherein they defined $k$ to be \emph{$c_0$-universal} if $k$ is bounded, $k(\cdot,x)\in C_0(X),\,\forall\,x\in X$ and its corresponding RKHS, $\eu{H}$ is dense in $C_0(X)$ w.r.t.~the uniform norm, where $X$ is a locally compact Hausdorff (LCH) space and $C_0(X)$ is the Banach space of bounded continuous functions vanishing at infinity, endowed with the uniform norm (see Section~\ref{Sec:Notation} for the definition of $C_0(X)$). 
\par As our contribution, in Section~\ref{subsec:main}, we present the following necessary and sufficient condition for a kernel to be \emph{$c_0$-universal}: $k$ is \emph{$c_0$-universal} if and only if the embedding in (\ref{Eq:signedmeasure}) is injective for $\mu\in M_b(X)$. It can be seen that this characterization naturally leads to understand the relation between \emph{$c_0$-universal} and \emph{characteristic} kernels, which is not straightforward with the characterization obtained by \citet[Theorem 2]{Carmeli-09}, wherein \emph{$c_0$-universality} is characterized by the injectivity of a certain integral operator on the space of square-integrable functions. Using this result, simple necessary and sufficient conditions are derived for translation invariant kernels on $\bb{R}^d$ (see Section~\ref{subsubsec:a1}), Fourier kernels on $\bb{T}^d$, the $d$-Torus (see Section~\ref{subsubsec:a2}) and radial kernels on $\bb{R}^d$ (see Section~\ref{subsubsec:a3}) to be \emph{$c_0$-universal}. Examples of \emph{$c_0$-universal} kernels on $\bb{R}^d$ include the Gaussian, Laplacian, $B_{2l+1}$-spline, inverse multiquadratics, Mat\'{e}rn class, etc.\vspace{2mm}
\par\noindent $c_b$-\textbf{universality:} 
The definition of \emph{$c_0$-universality} deals with $\eu{H}$ being dense in $C_0(X)$ w.r.t.~the uniform norm, where $X$ is an LCH space. Although the notion of \emph{$c_0$-universality} addresses limitations associated with both \emph{c-} and \emph{cc-universality}, it only approximates a subset of $C(X)$, i.e., it cannot deal with functions in $C(X)\backslash C_0(X)$. This limitation can be addressed by considering a larger class of functions to be approximated. 
\par To this end, we propose a notion of universality that is stronger than \emph{$c_0$-universality}: $k$ is said to be \emph{$c_b$-universal} if its corresponding RKHS, $\eu{H}$ is dense in $C_b(X)$, the space of bounded continuous functions on a topological space, $X$ (note that $C_0(X)\subset C_b(X)$). This notion of \emph{$c_b$-universality} is more applicable in learning theory than \emph{$c_0$-universality} as the target function, $f^\star$ can belong to $C_b(X)$ (which is a more natural assumption) instead of it being restrained to $C_0(X)$ (note that $C_0(X)$ only contains functions that vanish at infinity). We show in Section~\ref{subsec:main} that $k$ is \emph{$c_b$-universal} if and only if the embedding in (\ref{Eq:signedmeasure}) is injective for $\mu$ belonging to a \emph{certain} class of set functions (see Section~\ref{Sec:Notation} for the definition of set functions) defined on a normal topological space, $X$ (see Theorem~\ref{Thm:c0-universal} for details). Because of the technicalities involved in dealing with set functions, in this paper, we do not fully analyze this notion of universality unlike the other aforementioned notions, although it is an interesting problem to be resolved because of its applicability in learning theory. \vspace{2mm}
\par\noindent Based on the above discussion that relates injectivity of the embedding in (\ref{Eq:signedmeasure}) to various notions of universality, 
we also show how these notions of universality are related. If $X$ is compact, the notions of \emph{c-}, \emph{cc-}, \emph{$c_0$-} and \emph{$c_b$-universality} are equivalent. On the other hand, if $X$ is not compact, the notion of \emph{$c_0$-universality} is stronger than \emph{cc-universality}. I.e., if a kernel is \emph{$c_0$-universal}, then it is \emph{cc-universal} but not vice-versa (for example, the Gaussian kernel on $\bb{R}^d$ is shown to be \emph{$c_0$-universal} and therefore is \emph{cc-universal}, while the sinc kernel is \emph{cc-universal} but not \emph{$c_0$-universal}). 
We show in Section~\ref{subsubsec:a3} that the converse is true in the case of radial kernels on $\bb{R}^d$. Similarly, when $X$ is not compact (but an LCH space), the notion of \emph{$c_b$-universality} is stronger than \emph{$c_0$-universality}, and therefore \emph{cc-universality}. A summary of the relationship between various notions of universality is shown in Figure~\ref{Fig:results}.
\par To summarize our first contribution, we show that, by appropriately choosing $X$ and $\mu$ in (\ref{Eq:signedmeasure}), the injectivity of the embedding in (\ref{Eq:signedmeasure}) completely characterizes various notions of universality that are proposed in literature. Using this connection between universality and the injectivity of the embedding in (\ref{Eq:signedmeasure}), we relate all these notions of universality, which is summarized in Figure~\ref{Fig:results}.

\subsection{Contribution 2: Relation between characteristic and universal kernels}\label{subsec:relation}
\begin{figure*}
  \centering
  \begin{tabular}{cccc}
    \begin{minipage}{7.5cm}
      \center{\epsfxsize=6cm
      \epsffile{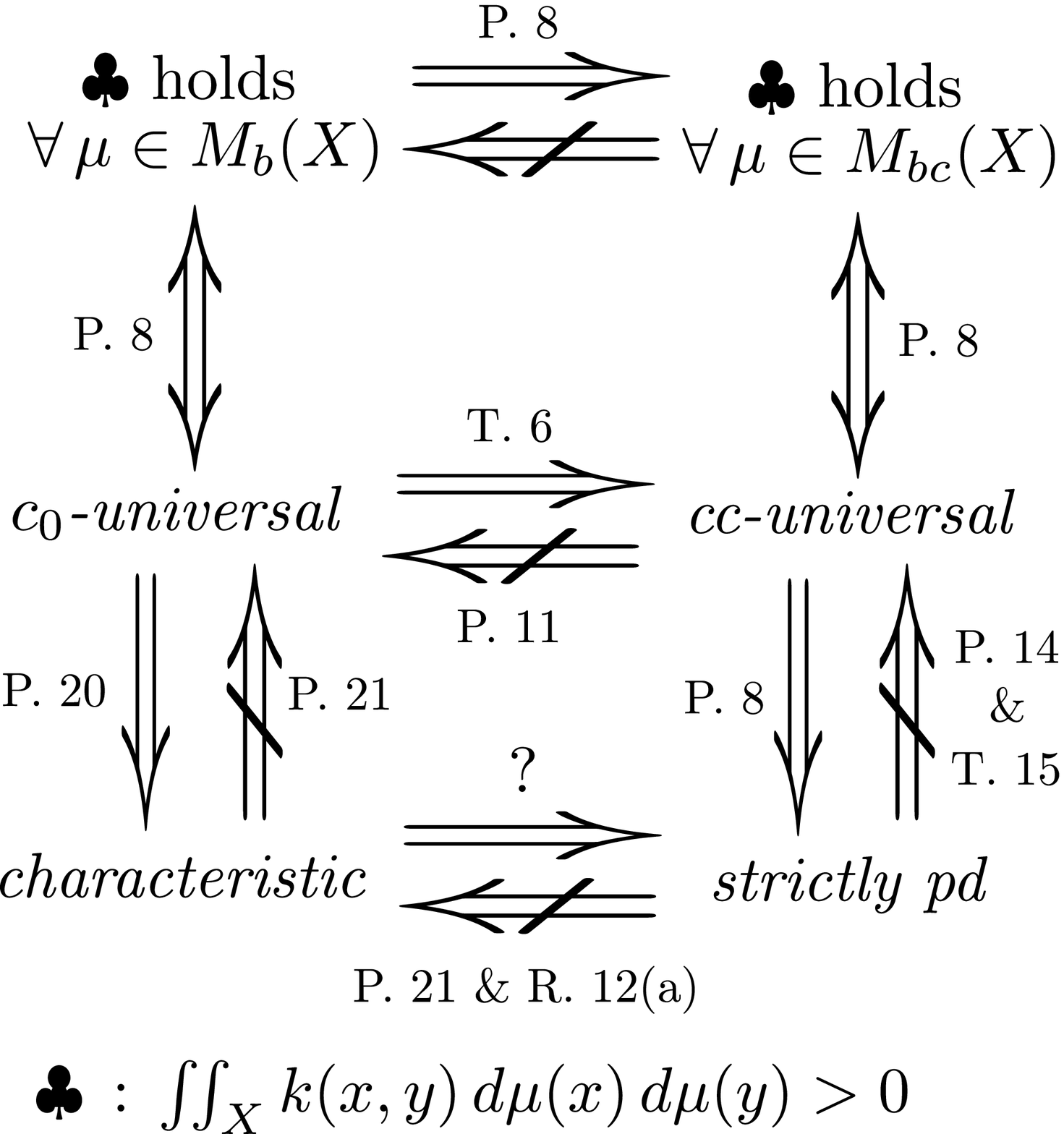}}\vspace{-2.5mm}
      {\small \center{(a)}}\hspace{2mm}
    \end{minipage}
    \begin{minipage}{7.5cm}
      \center{\epsfxsize=6cm
      \epsffile{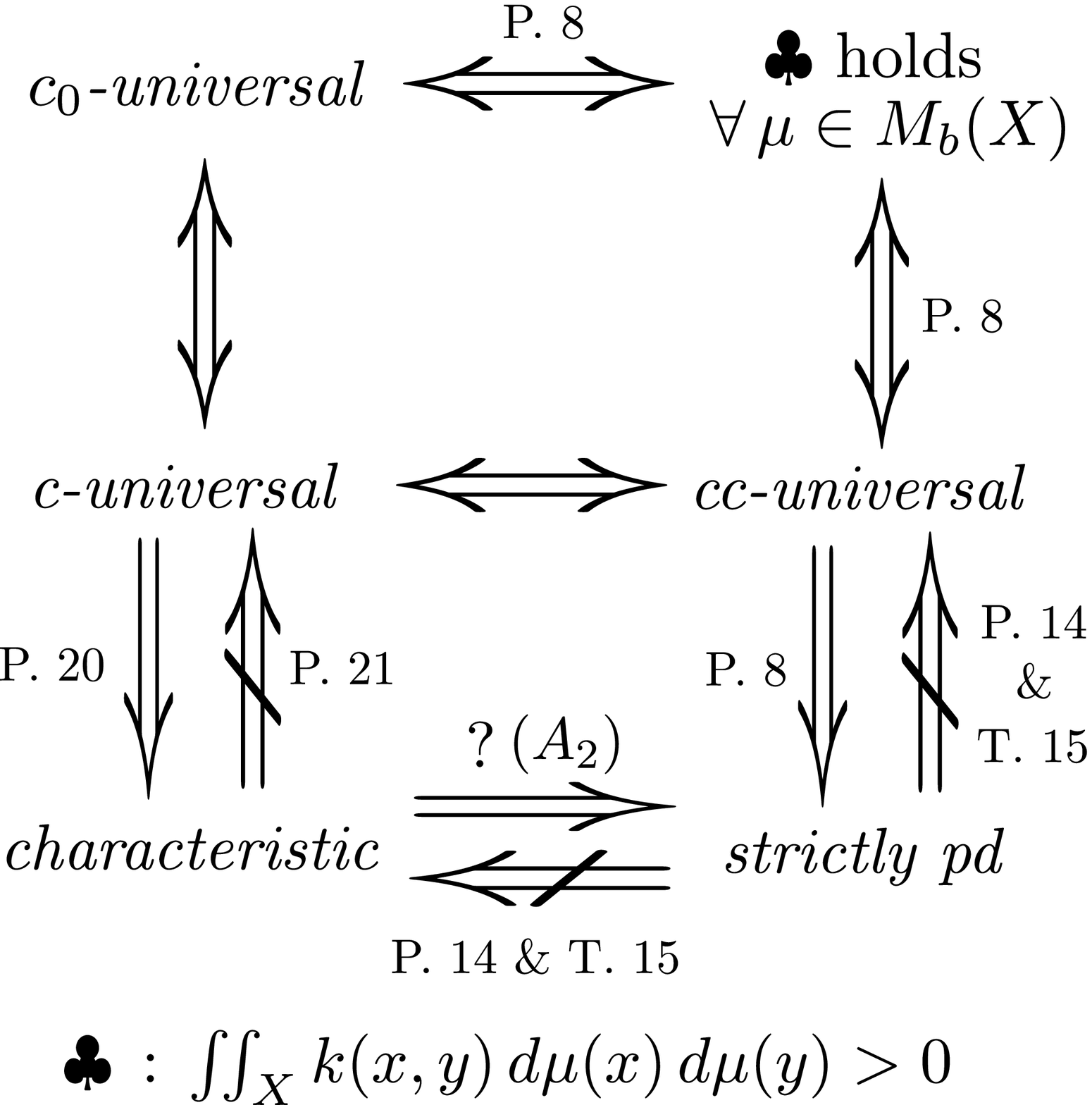}}\vspace{-2.5mm}
      {\small \center{(b)}}
    \end{minipage}
    \vspace{5mm}
  \end{tabular}
\begin{tabular}{cccc}
    \begin{minipage}{7.5cm}
      \center{\epsfxsize=6cm
      \epsffile{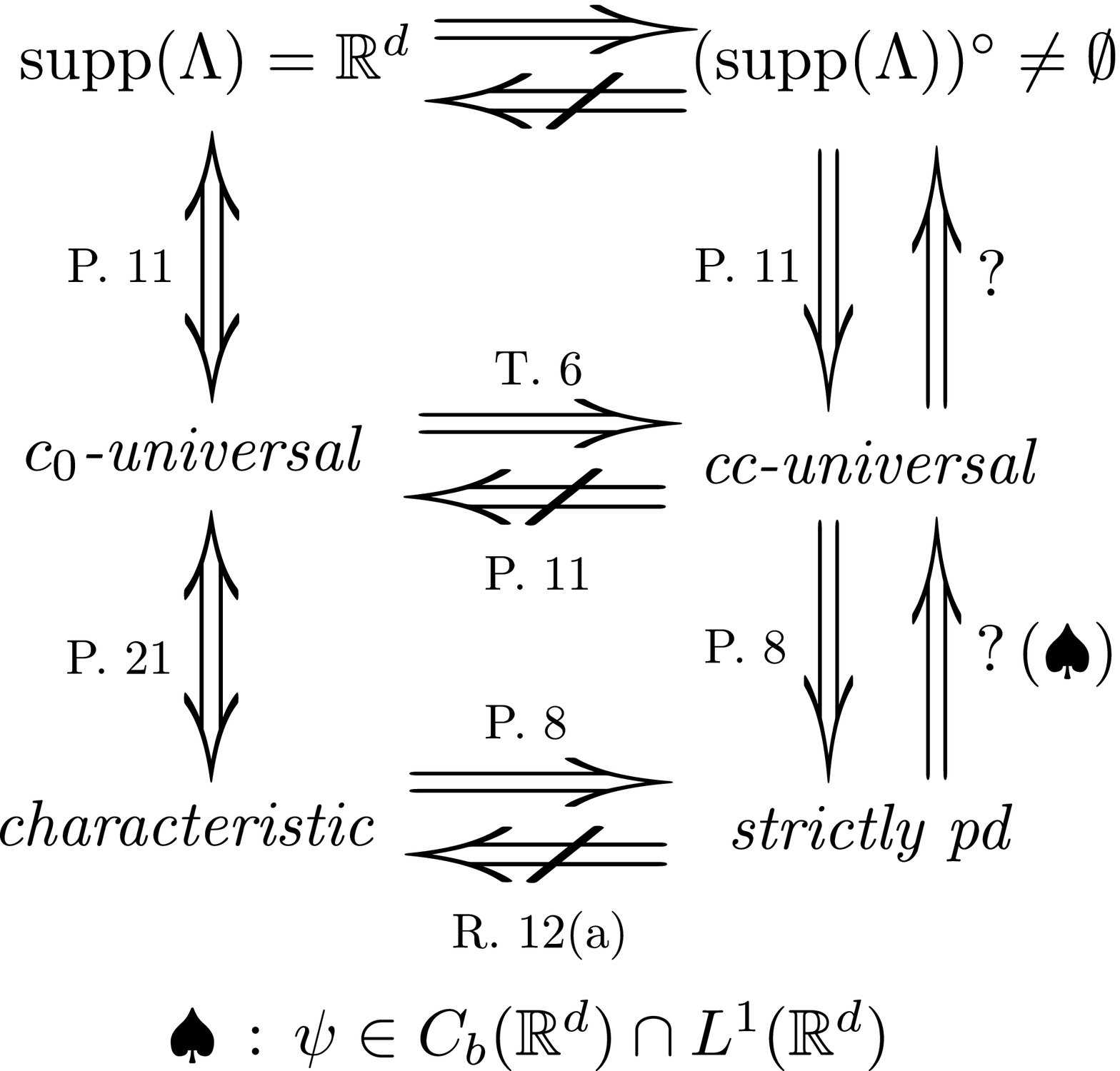}}\vspace{-2.5mm}
      {\small \center{(c)}}\hspace{2mm}
    \end{minipage}
    \begin{minipage}{7.5cm}
      \center{\epsfxsize=6cm
      \epsffile{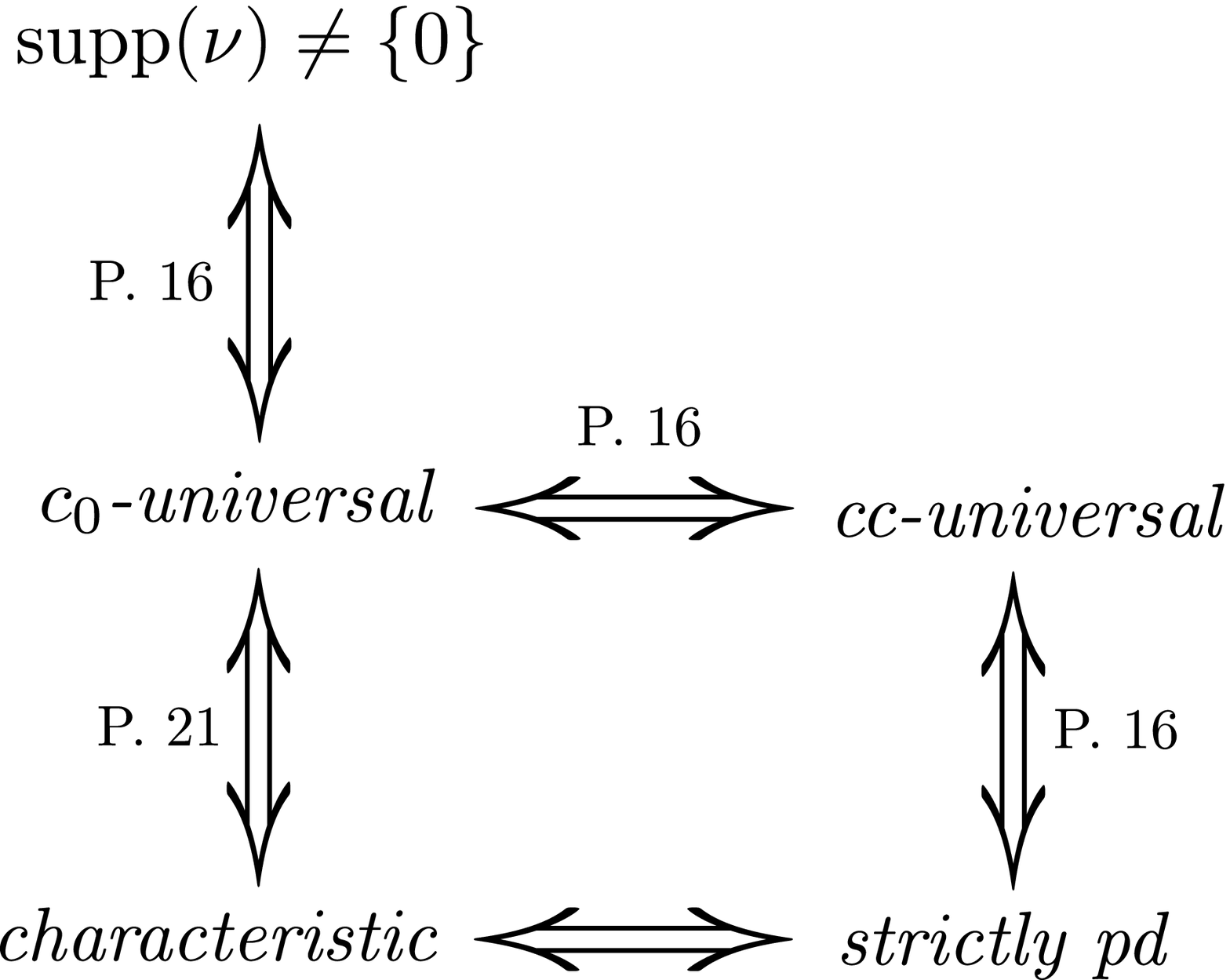}}\vspace{-2.5mm}
      {\small \center{(d)}}
    \end{minipage}
\vspace{-1mm}
  \end{tabular}
  \caption{Summary of results: The relationships between various notions are shown along with the reference. The letters ``P", ``R" and ``T" refer to Proposition, Remark and Theorem respectively. For example, P. 7 refers to Proposition 7. The implications which are open problems are shown with ``?". The trivial implications are shown without any reference. (a) $X$ is an LCH space. Refer to Section~\ref{Sec:Notation} for the definition of $M_b(X)$ and $M_{bc}(X)$. (b) The implications shown hold for any compact Hausdorff space, $X$. However, when $X=\bb{T}^d$, the $d$-Torus, with $k(x,y)=\psi((x-y)_{\text{mod}\,2\pi})$, where $\psi\in C(\bb{T}^d)$ is a positive definite (pd) function, the implication between \emph{characteristic} and \emph{strictly pd}, shown as ($A_2$) is valid, which follows from Proposition~\ref{pro:td} and Theorem~\ref{Thm:spd-sphere}. 
(c) $X=\bb{R}^d$ and $k(x,y)=\psi(x-y)$, where $\psi\in C_b(\bb{R}^d)$ is a pd function and the Fourier transform of a finite non-negative Borel measure, $\Lambda$ (see Theorem~\ref{Thm:Bochner} for details). If $\psi\in C_b(\bb{R}^d)\cap L^1(\bb{R}^d)$, then the implication shown as ($\spadesuit$) holds. Otherwise, it is not clear whether the implication holds. For a set $A$, $A^\circ$ represents its interior. (d) $X=\bb{R}^d$ and $k(x,y)=\varphi(\Vert x-y\Vert^2_2)$, where $\varphi$ is the Laplace transform of a finite non-negative Borel measure, $\nu$ on $[0,\infty)$ (see (\ref{Eq:schoenberg})).}
\label{Fig:results}
\vspace{-6mm}
\end{figure*}

\citet{Gretton-06} related universality and the characteristic property of $k$ by showing that if $k$ is \emph{c-universal}, then it is \emph{characteristic}. Besides this result, not much is known or understood about the relation between universal and characteristic kernels. In Section~\ref{subsec:main-char}, we relate universality and characteristic kernels by using the results in Section~\ref{subsec:main} that relate universality and the RKHS embedding of Radon measures. 
As an example, we show that a translation invariant kernel on $\bb{R}^d$ (in general, any locally compact Abelian group) or a radial kernel on $\bb{R}^d$ is \emph{$c_0$-universal} if and only if it is \emph{characteristic}. We also show that the converse to the result by \citet{Gretton-06} is not true, i.e., if a kernel is \emph{characteristic}, it need not be \emph{c-universal} \citep[see][Corollary 15]{Sriperumbudur-09a}. A summary of the relation between universal and characteristic kernels is shown in Figure~\ref{Fig:results}.
\par Using the embedding in (\ref{Eq:prob}), \citet{Gretton-06} proposed a metric, called the maximum mean discrepancy (MMD), on the space of all Borel probability measures, when $k$ is characteristic. One important theoretical question that is usually considered for metrics on probability measures is \cite[Chapter 11]{Dudley-02}: ``What is the nature of the topology induced by the probability metric in relation to the usual weak topology?" In probability theory, this question is important in understanding and proving central limit theorems. Although $k$ being characteristic is sufficient for MMD to be a metric, we show in Section~\ref{subsec:metric} that a notion stronger than the characteristic property is required to answer the above question. In particular, we show in Proposition~\ref{pro:weak} that if $X$ is an LCH space and $k$ is \emph{$c_0$-universal}, then the topology induced by MMD coincides with the usual weak topology on the space of Radon probability measures defined on $X$.\footnote{\citet{Sriperumbudur-09a} showed that if $X$ is a compact metric space and $k$ is \emph{c-universal}, then the topology induced by MMD coincides with the usual weak topology. The result for non-compact $X$ was left as an open question and is addressed in this paper, by applying the notion of \emph{$c_0$-universality}.} This result 
can be used to compare MMD to other probability metrics, such as the Dudley metric, total variation distance, Wasserstein distance, etc. We refer to \citet{Sriperumbudur-09a} for a detailed study on the comparison of MMD to other probability metrics.\vspace{2.5mm}

\par\noindent To summarize, our main contributions in this paper are: 
\begin{itemize}
\item[(a)] To establish the relationship between various notions of universality and the RKHS embedding, shown in (\ref{Eq:signedmeasure}), of finite signed Radon measures, and in turn present a novel measure embedding view point of universality compared to the classical function approximation view point.
\item[(b)] To clarify the relationship between universal and characteristic kernels.
\end{itemize} 
A summary of the results in this paper is shown in Figure~\ref{Fig:results}. In the following section, we introduce the notation and some definitions that are used throughout the paper. Supplementary results used in proofs are collected in Appendix A.

\section{Definitions \& Notation}\label{Sec:Notation}
Let $X$ be a topological space. $C(X)$ denotes the space of all continuous functions on $X$. $C_b(X)$ is the space of all bounded, continuous functions on $X$. For a locally compact Hausdorff space, $X$, $f\in C(X)$ is said to \emph{vanish at infinity} if for every $\epsilon>0$ the set $\{x:|f(x)|\ge\epsilon\}$ is compact. The class of all continuous $f$ on $X$ which vanish at infinity is denoted as $C_0(X)$. The spaces $C_b(X)$ and $C_0(X)$ are endowed with the uniform norm, $\Vert\cdot\Vert_u$ defined as $\Vert f\Vert_u:=\sup_{x\in X}|f(x)|$ for $f\in C_0(X)\subset C_b(X)$. 
\par If $Y$ denotes a topological vector space, we denote by $Y^\prime$ the vector space of continuous linear functionals on $Y$, and $Y^\prime$ is called the \emph{topological dual space} (in this paper, we simply refer to it as the \emph{dual}).
\par For a set $A$, we denote its interior as $A^\circ$.\vspace{2mm}\par\noindent
\textbf{Radon measure:} A signed \emph{Radon measure} $\mu$ on a Hausdorff space $X$ is a Borel measure on $X$ satisfying
\begin{itemize}
\item[$(i)$] $\mu(C)<\infty$ for each compact subset $C\subset X$,\vspace{-1mm}
\item[$(ii)$] $\mu(B)=\sup\{\mu(C)\,|\,C\subset B,\,C\,\text{compact}\}$ for each $B$ in the Borel $\sigma$-algebra of $X$.
\end{itemize}
$\mu$ is said to be finite if $\Vert\mu\Vert:=|\mu|(X)<\infty$, where $|\mu|$ is the total-variation of $\mu$. $M^b_+(X)$ denotes the space of all finite Radon measures on $X$ while $M_b(X)$ denotes the space of all finite signed Radon measures on $X$. The space of all Radon probability measures is denoted as $M^1_+(X):=\{\mu\in M^b_+(X):\mu(X)=1\}$. For $\mu\in M_b(X)$, the support of $\mu$ is defined as 
\begin{equation}\label{Eq:support}
\text{supp}(\mu)=\{x\in X\,\vert\,\text{for any open set}\,\,U\,\,\text{such that}\,\,x\in U,\,|\mu|(U)\ne 0\}.
\end{equation}
$M_{bc}(X)$ denotes the space of all compactly supported finite signed Radon measures on $X$. We refer the reader to \citet[Chapter 2]{Berg-84} for a general 
reference on the theory of Radon measures.\vspace{2mm}\par\noindent
\textbf{Finitely additive, regular set function:} A \emph{set function} is a function defined on a family of sets, and has values in $[-\infty,+\infty]$. 
\par A set function $\mu$ defined on a family $\tau$ of sets is said to be \emph{finitely additive} if $\emptyset\in\tau$, $\mu(\emptyset)=0$ and $\mu(\cup^n_{l=1}A_l)=\sum^n_{l=1}\mu(A_l)$, for every finite family $\{A_1,\ldots,A_n\}$ of disjoint subsets of $\tau$ such that $\cup^n_{l=1}A_l\in \tau$. 
\par A \emph{field of subsets} of a set $X$ is a non-empty family, $\Sigma$, of subsets of $X$ such that $\emptyset\in \Sigma$, $X\in\Sigma$, and for all $A,B\in \Sigma$, we have $A\cup B\in\Sigma$ and $B\backslash A\in\Sigma$. 
\par An additive set function $\mu$ defined on a field $\Sigma$ of subsets of a topological space $X$ is said to be \emph{regular} if for each $A\in\Sigma$ and $\epsilon>0$, there exists $B\in\Sigma$ whose closure is contained in $A$ and there exists $C\in\Sigma$ whose interior contains $A$ such that $|\mu(D)|<\epsilon$ for every $D\in\Sigma$ with $D:=C\backslash B$. \vspace{2mm}\par\noindent
\textbf{Positive definite (pd), strictly pd and conditionally strictly pd:} A function $k:X\times X\rightarrow\bb{R}$ is called \emph{positive definite} (pd) (\emph{resp.} conditionally pd) if, for all $n\in\bb{N}$ (\emph{resp.} $n\ge 2$), $\alpha_1,\ldots,\alpha_n\in\bb{R}$ (\emph{resp.} with $\sum^n_{j=1}\alpha_j=0$) and all $x_1,\ldots,x_n\in X$, we have 
\begin{equation}\label{Eq:pd}
\sum^n_{l,j=1}\alpha_l\alpha_jk(x_l	,x_j)\ge 0.
\end{equation}
Furthermore, $k$ is said to be \emph{strictly pd} (\emph{resp}. conditionally strictly pd) if, for mutually distinct $x_1,\ldots,x_n\in X$, equality in (\ref{Eq:pd}) only holds for $\alpha_1=\cdots=\alpha_n=0$.
\vspace{2mm}\par\noindent
\textbf{Fourier transform in $\bb{R}^d$:} For $X\subset\bb{R}^d$, let $L^p(X)$ denote the Banach space of $p$-power ($p\ge 1$) integrable functions w.r.t.~the Lebesgue measure. For $f\in L^1(\bb{R}^d)$, $\hat{f}$ and $\check{f}$ represent the Fourier transform and inverse Fourier transform of $f$ respectively, defined as 
\begin{eqnarray}
\hat{f}(y)&\!\!\!:=\!\!\!&(2\pi)^{-\frac{d}{2}}\int_{\bb{R}^d}e^{-iy^Tx} f(x)\,dx,\,\,y\in\bb{R}^d,\label{Eq:FT}\\
\check{f}(x)&\!\!\!:=\!\!\!&(2\pi)^{-\frac{d}{2}}\int_{\bb{R}^d}e^{ix^Ty} f(y)\,dy,\,\,x\in\bb{R}^d,\label{Eq:IFT}
\end{eqnarray}\par\noindent
where $i$ denotes the imaginary unit $\sqrt{-1}$. For a finite Borel measure, $\mu$ on $\bb{R}^d$, the Fourier transform of $\mu$ is given by
\begin{equation}
\hat{\mu}(\omega)=\int_{\bb{R}^d}e^{-i\omega^Tx}\,d\mu(x),\,\omega\in\bb{R}^d,\label{Eq:measure-FT}
\end{equation}
which is a bounded, uniformly continuous function on $\bb{R}^d$.
\vspace{2mm}\par\noindent
\textbf{Holomorphic and entire functions:} Let $D\subset\bb{C}^d$ be an open subset and $f:D\rightarrow\bb{C}$ be a function. $f$ is said to be \emph{holomorphic} at the point $z_0\in D$ if 
\begin{equation}f^\prime(z_0):=\lim_{z\rightarrow z_0} \frac{f(z_0)-f(z)}{z_0-z}\nonumber
s\end{equation} exists. Moreover, $f$ is called holomorphic if it is holomorphic at every $z_0\in D$. $f$ is called an \emph{entire function} if $f$ is holomorphic and $D=\bb{C}^d$.

\section{Characterization of Universal Kernels}\label{Sec:universality}
In Section~\ref{Sec:Introduction}, we have briefly discussed the relation between the embedding in (\ref{Eq:signedmeasure}) and various notions of universality. In Section~\ref{subsec:main}, we present and prove our main result (Theorem~\ref{Thm:c0-universal}), which relates universality and the embedding in (\ref{Eq:signedmeasure}). Theorem~\ref{Thm:c0-universal} shows that under appropriate assumptions on $\mu$ and $X$, the injectivity of the embedding in (\ref{Eq:signedmeasure}) is necessary and sufficient for a kernel to be \emph{c-}, \emph{cc-}, \emph{$c_0$-} or \emph{$c_b$-universal}. Using this result, it is shown that the notion of \emph{$c_0$-universality} is stronger than that of \emph{cc-universality}, i.e., if $k$ is \emph{$c_0$-universal}, then it is \emph{cc-universal} but not vice-versa. Then, in Proposition~\ref{pro:strictpd}, we obtain alternate necessary and sufficient conditions for the embedding in (\ref{Eq:signedmeasure}) to be injective, which resembles a condition for the kernel to be strictly pd (but not quite so!). However, in Proposition~\ref{pro:strictpd}, we show that strict positive definiteness of $k$ is a necessary condition for the embedding in (\ref{Eq:signedmeasure}) to be injective, i.e., for $k$ to be \emph{universal}. Using the characterization obtained in Proposition~\ref{pro:strictpd}, in Sections~\ref{subsubsec:a1}--\ref{subsubsec:a4}, we derive characterizations for universality that are easy to check, for specific classes of kernels, e.g., translation invariant kernels on $\bb{R}^d$ and $\bb{T}^d$, radial kernels on $\bb{R}^d$, Taylor-type kernels on $\bb{R}^d$, etc. The results of this section are summarized in Figure~\ref{Fig:results}. \par Before characterizing various notions of universality, let us revisit their formal definitions.
\begin{definition}[$c$-universal]\label{def:c-universal}
A continuous kernel $k$ on a compact Hausdorff space $X$ is called c-universal if the RKHS, $\eu{H}$ induced by $k$ is dense in $C(X)$ w.r.t.~the uniform norm, i.e., for every function $g\in C(X)$ and all $\epsilon> 0$, there exists an $f\in\eu{H}$ such that $\Vert f-g\Vert_u\le\epsilon$.
\end{definition}
\begin{definition}[$cc$-universal]\label{def:cc-universal}
A continuous kernel $k$ on a Hausdorff space $X$ is said to be cc-universal if the RKHS, $\eu{H}$ induced by $k$ is dense in $C(X)$ endowed with the topology of compact convergence, i.e., for any compact set $Z\subset X$, for any $g\in C(Z)$ and all $\epsilon> 0$, there exists an $f\in \eu{H}_{\vert Z}$ such that $\Vert f-g\Vert_u\le\epsilon$.
\end{definition}
\begin{definition}[$c_0$-universal]\label{def:c0-universal}
A bounded kernel, $k$ with $k(\cdot,x)\in C_0(X),\,\forall\,x\in X$ on a locally compact Hausdorff space, $X$ is said to be $c_0$-universal if the RKHS, $\eu{H}$ induced by $k$ is dense in $C_0(X)$ w.r.t.~the uniform norm, i.e., for every function $g\in C_0(X)$ and all $\epsilon>0$, there exists an $f\in \eu{H}$ such that $\Vert f-g\Vert_u\le\epsilon$.
\end{definition}
\begin{definition}[$c_b$-universal]\label{def:cb-universality}
A bounded continuous kernel, $k$ on a topological space, $X$, is said to be $c_b$-universal if the RKHS, $\eu{H}$ induced by $k$ is dense in $C_b(X)$ w.r.t.~the uniform norm, i.e., for any $g\in C_b(X)$ and all $\epsilon>0$, there exists an $f\in\eu{H}$ such that $\Vert f-g\Vert_u\le\epsilon$. 
\end{definition}
First note that the above definitions are valid only if $\eu{H}$ is included in the appropriate target space, i.e., $C(X)$ for \emph{c-} and \emph{cc-universality}, $C_0(X)$ for \emph{$c_0$-universality}, and $C_b(X)$ for \emph{$c_b$-universality}. By \citet[Lemma 4.28, Theorem 4.61]{Steinwart-08}, the assumptions made on the kernel in the above definitions ensure that the definitions are valid. 
Also note that all these definitions are equivalent when $X$ is compact as $C_0(X)=C_b(X)=C(X)$ for compact $X$. When $X$ is not compact, it is easy to see that \emph{$c_b$-universality} is stronger than \emph{$c_0$-universality}, i.e., if $k$ is \emph{$c_b$-universal}, then it is also \emph{$c_0$-universal}, but not vice-versa. On the other hand, it is not straightforward to see how the notions of \emph{cc-universal} and \emph{$c_0$-universal} are related when $X$ is non-compact. By characterizing \emph{$c_0$-universality} and \emph{cc-universality}, Theorem~\ref{Thm:c0-universal} in the following section, shows that the notion of \emph{$c_0$-universality} is stronger than \emph{cc-universality}, i.e., if a kernel is \emph{$c_0$-universal}, then it is \emph{cc-universal}, but not vice-versa. Based on these results, it follows that \emph{$c_b$-universality} is stronger than \emph{cc-universality} (but not vice-versa), when $X$ is non-compact.

\subsection{Main results}\label{subsec:main}
Before we state our main result, i.e., Theorem~\ref{Thm:c0-universal}, we need the following result, usually referred to as the Hahn-Banach theorem, which we quote from \citet[Theorem 3.5]{Rudin-91} (also see the remark following Theorem 3.5 in \citet{Rudin-91}).
\begin{theorem}[Hahn-Banach]\label{Thm:Hahn-Banach}
Suppose $A$ be a subspace of a locally convex topological vector space $Y$. Then $A$ is dense in $Y$ if and only if $A^\perp=\{0\}$, where
\begin{equation}\label{Eq:annihilator}
A^\perp:=\{T\in Y^\prime:\forall x\in A,\,\,T(x)=0\}.
\end{equation}
\end{theorem}
The following main result of this paper, which presents a necessary and sufficient condition for $k$ to be \emph{c-}, \emph{cc-}, \emph{$c_0$-} or \emph{$c_b$-universal}. hinges on the above theorem, where we choose $A$ to be the RKHS, $\eu{H}$ and $Y$ to be $C(X)$, $C_0(X)$ or $C_b(X)$ for which $Y^\prime$ is known through the Riesz representation theorem.
\begin{theorem}[Characterization of universal kernels]\label{Thm:c0-universal}
The following hold:
\begin{itemize}
\item[(a)] Let $X$ be a compact Hausdorff space with $k$ being continuous. Then $k$ is c-universal if and only if the embedding,
\begin{equation}\label{Eq:c-embedding}
\mu\mapsto \int_X k(\cdot,x)\,d\mu(x),\,\,\mu\in M_b(X),
\end{equation}
is injective.
\item[(b)] Let $X$ be an LCH space and $k\in C_b(X\times X)$. Then $k$ is cc-universal if and only if the embedding,
\begin{equation}\label{Eq:cc-universal}
\mu\mapsto\int_Xk(\cdot,x)\,d\mu(x),\,\,\mu\in M_{bc}(X),
\end{equation}
is injective.
\item[(c)] Let $X$ be an LCH space with the kernel, $k$ being bounded and $k(\cdot,x)\in C_0(X),\,\forall\,x\in X$. Then $k$ is $c_0$-universal if and only if the embedding,
\begin{equation}\label{Eq:c0-embedding}
\mu\mapsto \int_X k(\cdot,x)\,d\mu(x),\,\,\mu\in M_b(X),
\end{equation}
is injective.
\item[(d)] Let $X$ be a normal topological space and let $M_{rba}(X)$ be the space of all finitely additive, regular, bounded set functions defined on the field generated by the closed sets of $X$. Then, a bounded continuous kernel, $k$ is $c_b$-universal if and only if the embedding,
\begin{equation}\label{Eq:cb}
\mu\mapsto\int_X k(\cdot,x)\,d\mu(x),\,\mu\in M_{rba}(X),
\end{equation}
is injective.
\end{itemize}
\end{theorem}
\begin{proof}
First, we prove $(c)$, from which $(a)$ follows.\vspace{1mm}\\ 
$(c)$ By Definition~\ref{def:c0-universal}, $k$ is \emph{$c_0$-universal} if $\eu{H}$ is dense in $C_0(X)$. We now invoke Theorem~\ref{Thm:Hahn-Banach} to characterize the denseness of $\eu{H}$ in $C_0(X)$, which means we need to consider the dual $C^\prime_0(X):=(C_0(X))^\prime$ of $C_0(X)$. By the Riesz representation theorem \cite[Theorem 7.17]{Folland-99}, $C^\prime_0(X)=M_b(X)$ in the sense that there is a bijective linear isometry $\mu\mapsto T_\mu$ from $M_b(X)$ onto $C^\prime_0(X)$, given by the natural mapping, 
\begin{equation}
T_\mu(f)=\int_X f\, d\mu,\,\,f\in C_0(X).
\end{equation}
Therefore, by Theorem~\ref{Thm:Hahn-Banach}, $\eu{H}$ is dense in $C_0(X)$ if and only if 
\begin{equation}\label{Eq:annihilator-h}
\eu{H}^\perp:=\left\{\mu\in M_b(X):\forall\,f\in\eu{H},\,\,\int_X\,f\,d\mu=0\right\}=\{0\}.
\end{equation}
($\,\Leftarrow\,$) Suppose (\ref{Eq:c0-embedding}) is injective, i.e., for $\mu\in M_b(X)$, $\int_X k(\cdot,x)\,d\mu(x)=0\Rightarrow \mu=0$. Then by Lemma~\ref{lem:interchange} (see Appendix A), we have 
\begin{equation}
\int_X f\, d\mu=\Big\langle f,\int_X k(\cdot,x)\,d\mu(x)\Big\rangle_\eu{H}=0,\,\forall\,f\in\eu{H}\Rightarrow\mu=0,\nonumber
\end{equation}
which by (\ref{Eq:annihilator-h}) means $\eu{H}$ is dense in $C_0(X)$ and therefore $k$ is \emph{$c_0$-universal}.\vspace{1.5mm}\\
($\,\Rightarrow\,$) We need to prove that if $\eu{H}$ is dense in $C_0(X)$ then $(\int_X k(\cdot,x)\,d\mu(x)=0\Rightarrow\mu=0)$ holds. This is equivalent to showing that if $(\int_X k(\cdot,x)\,d\mu(x)=0\Rightarrow\mu=0)$ does not hold, then $\eu{H}$ is not dense in $C_0(X)$. Suppose $(\int_X k(\cdot,x)\,d\mu(x)=0\Rightarrow\mu=0)$ does not hold, i.e., $\exists\,0\ne\mu\in M_b(X)$ such that $\int_X k(\cdot,x)\,d\mu(x)=0$, which means $\exists\,0\ne\mu\in M_b(X)$ such that $\int_X f\,d\mu=0$ for every $f\in\eu{H}$, then,
by (\ref{Eq:annihilator-h}), $\eu{H}$ is not dense in $C_0(X)$.\vspace{2mm}\\
$(a)$ When $X$ is compact, $C_0(X)$ coincides with $C(X)$, which means \emph{c-universality} and \emph{$c_0$-universality} are equivalent. Therefore, $k$ is \emph{c-universal} if and only if the embedding in (\ref{Eq:c-embedding}) is injective.\vspace{2mm}\\
$(b)$ The proof is similar to that of $(a)$ except that we need to consider the dual of $C(X)$ endowed with the topology of compact convergence (a locally convex topological vector space) to characterize the denseness of $\eu{H}$ in $C(X)$. It is known \citep{Hewitt-50} that $C^\prime(X)=M_{bc}(X)$ in the sense that there is a bijective linear isometry $\mu\mapsto T_\mu$ from $M_{bc}(X)$ onto $C^\prime(X)$, given by the natural mapping, $T_\mu(f)=\int_X f\, d\mu,\,\,f\in C(X)$. The rest of the proof is verbatim with $M_b(X)$ replaced by $M_{bc}(X)$.\vspace{2mm}\\
$(d)$ The proof is very similar to that of $(a)$ , wherein we identify $(C_b(X))^\prime \cong M_{rba}(X)$ such that $T\in (C_b(X))^\prime$ and $\mu\in M_{rba}(X)$ satisfy $T(f)=\int_X f\,d\mu,\,f\in C_b(X)$ \citep[p. 262]{Dunford-58}. Here, $\cong$ represents the isometric isomorphism. The rest of the proof is verbatim with $M_b(X)$ replaced by $M_{rba}(X)$.
\end{proof}
Theorem~\ref{Thm:c0-universal} can also be interpreted as: for appropriate assumptions on $X$ and $\mu$, the embedding in (\ref{Eq:signedmeasure}) is injective if and only if the kernel is \emph{universal}, therefore relating universality and injective RKHS embedding of finite signed Radon measures. In other words, Theorem~\ref{Thm:c0-universal} provides a novel measure embedding view point of universality compared to its well-known function approximation view point. Based on Theorem~\ref{Thm:c0-universal}, the following remarks can be made.
\begin{remark}\label{rem:universal}
(a) Theorem~\ref{Thm:c0-universal} provides a necessary and sufficient condition for c-universality~---~$k$ is c-universal if and only if the embedding in (\ref{Eq:c-embedding}) is injective~---~while \citet{Steinwart-01} provided only a sufficient condition (in terms of the feature maps being an algebra; see \citet[Theorem 4.56]{Steinwart-08} for details) using the Stone-Weierstra\ss\;\,theorem. Therefore, Theorem~\ref{Thm:c0-universal} differs from and generalizes the result by \citet{Steinwart-01}.\vspace{2mm}\\
(b) Note that the embedding in (\ref{Eq:cc-universal}) is injective if and only if for any compact set $Z\subset X$, the embedding 
\begin{equation}\label{Eq:compact-michelli}
\mu\mapsto\int_Z k(\cdot,x)\,d\mu(x),\,\mu\in M_b(Z),
\end{equation}
is injective. \citet[Proposition 1]{Micchelli-06} have shown that for any compact set $Z\subset X$, the embedding in (\ref{Eq:compact-michelli}) is injective if and only if the set $K(Z)=\overline{\emph{span}}\{k(\cdot,y)\,:\,y\in Z\}$ is dense in $C(Z)$ w.r.t.~the uniform norm. Therefore, it is clear that $k$ is cc-universal if and only if it is universal in the sense of \citet{Micchelli-06}. See also \citet[Remark 1]{Carmeli-09}.
\vspace{2mm}\\
(c) By comparing the embeddings in (\ref{Eq:cc-universal}) and (\ref{Eq:c0-embedding}), since $M_{bc}(X)\subset M_b(X)$, it is clear that $c_0$-universality is stronger than cc-universality, i.e., if a kernel is $c_0$-universal (satisfies (\ref{Eq:c0-embedding})), then it is cc-universal (satisfies (\ref{Eq:cc-universal})). In general, the converse is not true (see Proposition~\ref{pro:rd} and Example~\ref{Ex:c0-universal}). However, we will show these notions to be equivalent in the case of radial kernels on $\bb{R}^d$ (see Proposition~\ref{pro:a3}).\vspace{2mm}\\
(d) \citet[Theorems 2,4]{Carmeli-09} provided characterizations for $c_0$- and cc-universality in terms of the injectivity of an integral operator on the space of square-integrable functions, whereas our characterizations in Theorem~\ref{Thm:c0-universal} deal with the injectivity of an embedding that maps finite signed Radon measures into an RKHS, $\eu{H}$. Since the latter
can be seen as a generalization of the embedding in (\ref{Eq:prob}) that deals with characteristic kernels, 
our characterizations can be used in a straightforward way to relate universal and characteristic kernels (see Section~\ref{Sec:characteristic} for details).\vspace{2mm}\\
(e) Note that $M_{rba}(X)$ in (\ref{Eq:cb}) does not contain any measure~---~though a set function in $M_{rba}(X)$ can be extended to a measure~---as measures are countably additive and defined on a $\sigma$-field. Since $\mu$ in Theorem~\ref{Thm:c0-universal}(d) is not a measure but a finitely additive set function defined on a field, it is not clear how to deal with the integral in (\ref{Eq:cb}). Because of the technicalities involved in dealing with set functions, we do not further pursue the notion of $c_b$-universality in this paper.\vspace{-1mm}
\end{remark}
Based on Theorem~\ref{Thm:c0-universal}, the following result provides an alternate and equivalent characterization of \emph{universality} or injectivity of the embedding in (\ref{Eq:signedmeasure}), which is easier to interpret, as it resembles the condition of $k$ being strictly pd (though not quite exactly the same). This alternate characterization is then used in Sections~\ref{subsubsec:a1}--\ref{subsubsec:a3} to obtain easily checkable conditions for the universality of specific classes of kernels. We also show that strictly pd is a necessary condition for \emph{universality}.
\begin{proposition}\label{pro:strictpd}
Suppose the assumptions in Theorem~\ref{Thm:c0-universal} hold. Then,
\begin{itemize}
\item[(a)] $k$ is $c$-universal if and only if
\begin{equation}\label{Eq:strictpd-2}
\int\!\!\int_X k(x,y)\,d\mu(x)\,d\mu(y)>0,\,\forall\,0\ne\mu\in M_b(X).
\end{equation}
\item[(b)] $k$ is cc-universal if and only if
\begin{equation}\label{Eq:strictpd-1}
\int\!\!\int_X k(x,y)\,d\mu(x)\,d\mu(y)>0,\,\forall\,0\ne\mu\in M_{bc}(X).
\end{equation}
\item[(c)] $k$ is $c_0$-universal if and only if
\begin{equation}\label{Eq:strictpd}
\int\!\!\int_X k(x,y)\,d\mu(x)\,d\mu(y)>0,\,\forall\,0\ne\mu\in M_b(X).
\end{equation}
\item[(d)] If $k$ is c-, cc- or $c_0$-universal, then it is strictly pd.
\end{itemize}
\end{proposition}
\begin{proof}
We only prove $(c)$. The proof of $(b)$ is exactly the same as that of $(c)$ with $M_b(X)$ replaced by $M_{bc}(X)$, while the proof of $(a)$ is trivial. \vspace{1mm}\\
$(c)$ ($\,\Leftarrow\,$) Suppose $k$ is not \emph{$c_0$-universal}. By Theorem~\ref{Thm:c0-universal}(c), there exists $0\ne\mu\in M_b(X)$ such that $\int_X k(\cdot,x)\,d\mu(x)=0$, which implies $\Vert\int_X k(\cdot,x)\,d\mu(x)\Vert_\eu{H}=0$. This means 
\begin{equation}
0=\Big\langle\int_X k(\cdot,x)\,d\mu(x),\int_X k(\cdot,x)\,d\mu(x)\Big\rangle_\eu{H}
\stackrel{(e)}{=}\int\int k(x,y)\,d\mu(x)\,d\mu(y),\nonumber
\end{equation}
where $(e)$ follows from Lemma~\ref{lem:interchange} (see Appendix A). By our assumption in (\ref{Eq:strictpd}), this leads to a contradiction. Therefore, if (\ref{Eq:strictpd}) holds, then $k$ is \emph{$c_0$-universal}.\vspace{2mm}\\
($\,\Rightarrow\,$) Suppose there exists $0\ne\mu\in M_b(X)$ such that $\int\!\!\int_X k(x,y)\,d\mu(x)\,d\mu(y)=0$, i.e., $\Vert\int_X k(\cdot,x)\,d\mu(x)\Vert_\eu{H}=0$, which implies $\int_X k(\cdot,x)\,d\mu(x)=0$. Therefore, the embedding in (\ref{Eq:c0-embedding}) is not injective, which by Theorem~\ref{Thm:c0-universal} implies that $k$ is not \emph{$c_0$-universal}. Therefore, if $k$ is \emph{$c_0$-universal}, then $k$ satisfies (\ref{Eq:strictpd}).\vspace{2mm}\\
$(d)$ Suppose $k$ is not strictly pd. This means for some $n\in\bb{N}$ and for  mutually distinct $x_1,\ldots,x_n\in X$, there exists $\bb{R}\ni\alpha_j\ne 0$ for some $j\in\{1,\ldots,n\}$ such that 
\begin{equation}\label{Eq:spdk}
\sum^n_{l,j=1}\alpha_l\alpha_jk(x_l,x_j)=0.
\end{equation} 
Define $\mu:=\sum^n_{j=1}\alpha_j\delta_{x_j}$, where $\delta_x$ represents the Dirac measures at $x$. Clearly $\mu\ne 0$ and $\mu\in M_{bc}(X)$. From (\ref{Eq:spdk}), it is clear that
$\int\!\!\int_X k(x,y)\,d\mu(x)\,d\mu(y)=0$. Therefore, by Proposition~\ref{pro:strictpd}(b), $k$ is not \emph{cc-universal}. The result for \emph{$c_0$-universality} follows from Remark~\ref{rem:universal}(c), while the result for \emph{c-universality} is trivial. See \citet[Corollary 5]{Carmeli-09}, \citet[Proposition 4.54, Example 4.11]{Steinwart-08} and \citet[Footnote 4]{Sriperumbudur-09a}.\vspace{-6mm}
\end{proof}
\begin{remark}\label{rem:strictlypd}
(a) Although the conditions in (\ref{Eq:strictpd-2})-(\ref{Eq:strictpd}) resemble the strictly pd condition, they are not equivalent. By combining any of (a)-(c) with (d) in Proposition~\ref{pro:strictpd}, it is easy to see that if $k$ satisfies any of (\ref{Eq:strictpd-2})-(\ref{Eq:strictpd}), then it is strictly pd. However, the converse is not true (see Remark~\ref{rem:Rd}(a) and the discussion following Example \ref{Exm:td}; also refer to \citet[Proposition 4.60, Theorem 4.62]{Steinwart-08} for the related discussion). We show in Section~\ref{subsubsec:a3} that in the case of radial kernels on $\bb{R}^d$, the converse is true, i.e., $k$  being strictly pd is also sufficient for it to be cc- or $c_0$-universal (see Proposition~\ref{pro:a3}).\vspace{2mm}\\
(b) The condition on $k$ in (\ref{Eq:strictpd}) can be seen as a generalization of integrally strictly pd kernels \citep[Section 6]{Stewart-76}: $\int\!\!\int_X k(x,y)f(x)f(y)\,dx\,dy>0$ for all $f\in L^2(\bb{R}^d)$, which is the strictly positive definiteness of the integral operator given by the kernel.\vspace{-1mm}
\end{remark}
A summary of results based on Theorem~\ref{Thm:c0-universal}, Remarks~\ref{rem:universal}, \ref{rem:strictlypd} and Proposition~\ref{pro:strictpd} is shown in Figures~\ref{Fig:results}(a) and \ref{Fig:results}(b). 
\par Although the conditions in (\ref{Eq:strictpd-2})-(\ref{Eq:strictpd}) are easy to interpret, they are not always easy to check. To this end, in the remainder of this section, we present easily checkable characterizations for the following classes of kernels. These classes of kernels are both mathematically and practically interesting as many of the popular kernels used in machine learning, e.g., Gaussian, Laplacian, exponential, etc., fall in these classes (see Examples~\ref{Ex:c0-universal}--\ref{Exm:radial} for more examples). 
\begin{itemize}
\item[($A_1$)] $k$ is translation invariant on $\bb{R}^d\times \bb{R}^d$, i.e., $k(x,y)=\psi(x-y)$, where $0\ne\psi\in C_b(\bb{R}^d)$ is a pd function on $\bb{R}^d$.\footnote{$\psi$ is said to be a pd function on $\bb{R}^d$ if $k(x,y)=\psi(x-y)$ is pd.}
\item[($A_2$)] \emph{Fourier kernel:} $k$ is translation invariant on $\bb{T}^d\times \bb{T}^d$, where $\bb{T}^d:=[0,2\pi)^d$, the $d$-Torus, i.e., $k(x,y)=\psi((x-y)_{\text{mod}\,2\pi})$, where $\psi\in C(\bb{T}^d)$ is a pd function on $\bb{T}^d$.
\item[($A_3$)] $k$ is a radial kernel on $\bb{R}^d\times \bb{R}^d$, i.e., there exists a finite nonnegative Borel measure, $\nu$ on $[0,\infty)$ such that for all $x,y\in\bb{R}^d$, 
\begin{equation}\label{Eq:schoenberg}
k(x,y)=\int_{[0,\infty)}e^{-t\Vert x-y\Vert^2_2}\,d\nu(t).
\end{equation}
These kernels are also called \emph{Schoenberg kernels} \citep[Corollary 7.12, Theorem 7.13]{Wendland-05}.\footnote{Note that $k$ is a scale mixture of Gaussian kernels.}
\item[($A_4$)] $X$ is an LCH space with bounded $k$. Let $k(x,y)=\sum_{j\in I}\phi_j(x)\phi_j(y),\,\,(x,y)\in X\times X$, where we assume the series converges uniformly on $X\times X$. $\{\phi_j:j\in I\}$ is a set of continuous real-valued functions on $X$ where $I$ is a countable index set.
\end{itemize}

\subsection{Translation invariant kernels on $\bb{R}^d$: ($A_1$)}\label{subsubsec:a1}
The following result provides an easily checkable characterization for $k$ to be \emph{$c_0$-universal} or \emph{cc-universal} (we do not consider \emph{c-universality} as $X=\bb{R}^d$ is not compact) when $k$ is translation invariant on $\bb{R}^d$, i.e., when $k$ satisfies ($A_1$). Before we present the result, we need a theorem due to Bochner that characterizes translation invariant kernels on $\bb{R}^d$, which is quoted from \citet[Theorem 6.6]{Wendland-05}.
\begin{theorem}[Bochner]\label{Thm:Bochner}
$\psi\in C_b(\bb{R}^d)$ is pd on $\bb{R}^d$ if and only if it is the Fourier transform of a finite nonnegative Borel measure $\Lambda$ on $\bb{R}^d$, i.e.,
\begin{equation}\label{Eq:bochner}
\psi(x)=\int_{\bb{R}^d}e^{-ix^T\omega}\,d\Lambda(\omega),\,x\in\bb{R}^d.
\end{equation}
\end{theorem}
\begin{proposition}[Translation invariant kernels on $\bb{R}^d$]\label{pro:rd}
Suppose \emph{($A_1$)} holds.
\begin{itemize}
\item[(a)] Let $\psi \in C_0(\bb{R}^d)$. Then $k$ is $c_0$-universal if and only if $\emph{supp}(\Lambda)=\bb{R}^d$.\footnote{See (\ref{Eq:support}) for the definition of support of a Borel measure.}
\item[(b)] If $\emph{supp}(\psi)$ is compact, then $k$ is $c_0$-universal.
\item[(c)] If $(\emph{supp}(\Lambda))^\circ\ne\emptyset$, then $k$ is cc-universal.
\end{itemize}
\end{proposition}
\begin{proof}
$(a)$ ($\,\Leftarrow\,$) Consider $\int\!\!\int_{\bb{R}^d} k(x,y)\,d\mu(x)\,d\mu(y)$ for any $0\ne\mu\in M_b(\bb{R}^d)$ with $k(x,y)=\psi(x-y)$.
\begin{eqnarray}
B:=\int\!\!\int_{\bb{R}^d} k(x,y)\,d\mu(x)\,d\mu(y)&\!\!=\!\!&\int\!\!\int_{\bb{R}^d}\psi(x-y)\,d\mu(x)\,d\mu(y)\nonumber\\
&\!\!\stackrel{(d)}{=}\!\!&\int\!\!\int\!\!\int_{\bb{R}^d}e^{-i(x-y)^T\omega}\,d\Lambda(\omega)\,d\mu(x)\,d\mu(y)\nonumber\\
&\!\!\stackrel{(e)}{=}\!\!&\int\!\!\int_{\bb{R}^d}e^{-ix^T\omega}\,d\mu(x)\int_{\bb{R}^d}e^{iy^T\omega}\,d\mu(y)\,d\Lambda(\omega)\nonumber
\end{eqnarray}
\begin{eqnarray}
&\!\!\stackrel{(f)}{=}\!\!&\int_{\bb{R}^d}\hat{\mu}(\omega)\overline{\hat{\mu}(\omega)}\,d\Lambda(\omega)\nonumber\\
&\!\!=\!\!&\int_{\bb{R}^d}\left|\hat{\mu}(\omega)\right|^2\,d\Lambda(\omega),\label{Eq:translation}
\end{eqnarray}
where Theorem~\ref{Thm:Bochner} is invoked in $(d)$, Fubini's theorem~\cite[Theorem 2.37]{Folland-99} in $(e)$ and (\ref{Eq:FT}) in $(f)$. If $\text{supp}(\Lambda)=\bb{R}^d$, then it is clear that $B>0$. Therefore, by Proposition~\ref{pro:strictpd}(c), $k$ is \emph{$c_0$-universal.}\vspace{1mm}\\
($\Rightarrow$) Suppose $k$ is \emph{$c_0$-universal}, which by Theorem~\ref{Thm:c0-universal}(a) means that $\mu\mapsto \int_{\bb{R}^d} k(\cdot,x)\,d\mu(x)$ is injective for $\mu\in M_b(\bb{R}^d)$. This means $\mu\mapsto \int_{\bb{R}^d} k(\cdot,x)\,d\mu(x)$ is injective for $\mu\in M^1_+(\bb{R}^d)$ and therefore Theorem 7 in \citet{Sriperumbudur-08} 
yields $\text{supp}(\Lambda)=\bb{R}^d$.\vspace{2mm}\\
$(b)$ The proof is the same as that of Corollary 10 in \citet{Sriperumbudur-09a}. Since $\text{supp}(\psi)$ is compact in $\bb{R}^d$, by the Paley-Wiener theorem \citep[Theorem 7.23]{Rudin-91}, we deduce that $\text{supp}(\Lambda)=\bb{R}^d$. Therefore, the result follows from Proposition~\ref{pro:rd}(a).\vspace{2mm}\\
$(c)$ Consider $\int\!\!\int_{\bb{R}^d} k(x,y)\,d\mu(x)\,d\mu(y)$ with $k(x,y)=\psi(x-y)$ and $\mu\in M_{bc}(\bb{R}^d)$. Since (\ref{Eq:translation}) holds for any $\mu\in M_b(\bb{R}^d)$, it also holds for any $\mu\in M_{bc}(\bb{R}^d)$, i.e.,
\begin{equation}
B:=\int\!\!\int_{\bb{R}^d} k(x,y)\,d\mu(x)\,d\mu(y)=\int_{\bb{R}^d}|\hat{\mu}(\omega)|^2\,d\Lambda(\omega).\nonumber
\end{equation}
Since $\mu\in M_{bc}(\bb{R}^d)$, by the Paley-Wiener theorem \citep[Theorem 7.23]{Rudin-91}, we obtain that $\hat{\mu}$ cannot vanish over an open set in $\bb{R}^d$ and $\text{supp}(\hat{\mu})=\bb{R}^d$. Therefore if $(\text{supp}(\Lambda))^\circ\ne\emptyset$, then $B>0$ for every $0\ne\mu\in M_{bc}(\bb{R}^d)$ and the result follows from Proposition~\ref{pro:strictpd}(b).
\end{proof}
Proposition~\ref{pro:rd} can easily be extended to locally compact Abelian groups by using the ideas in \citet{Fukumizu-08b}. Note that Proposition~\ref{pro:rd}(c) matches with Proposition 15 in \citet{Micchelli-06}, which is not surprising (see Remark~\ref{rem:universal}(b)). Based on Proposition~\ref{pro:rd}, in the following, we provide some examples of \emph{$c_0$}- and \emph{cc-universal} kernels that are translation invariant kernels on $\bb{R}^d$.
\begin{example}\label{Ex:c0-universal}
Let $d\Lambda(\omega)=(2\pi)^{-d/2}\hat{\psi}(\omega)\,d\omega$. Note that $\emph{supp}(\Lambda)=\emph{supp}(\hat{\psi})$. The following kernels satisfy $\emph{supp}(\hat{\psi})=\bb{R}^d$ and therefore are both $c_0$- and cc-universal.
\begin{enumerate}
\item[(1)] Gaussian, $\psi(x)=\exp\left(-\frac{\Vert x\Vert^2_2}{2\sigma^2}\right),\,\sigma>0$ with $\hat{\psi}(\omega)=\sigma^d\exp\left(-\frac{\sigma^2 \Vert \omega\Vert^2_2}{2}\right)$.   
\item[(2)] Laplacian, $\psi(x)=\exp\left(-\sigma\Vert x\Vert_1\right),\,\sigma>0$ with $\hat{\psi}(\omega)=\left(\frac{2}{\pi}\right)^{d/2}\prod^d_{j=1}\frac{\sigma}{\sigma^2+\omega^2_j}$, where $\omega=(\omega_1,\ldots,\omega_d)$.
\item[(3)] $B_1$-spline, $\psi(x)=\prod^d_{j=1}(1-|x_j|)\mathds{1}_{[-1,1]}(x_j)$ with $\hat{\psi}(\omega)=\prod^d_{j=1}\frac{4}{\sqrt{2\pi}}\frac{\sin^2(\omega_j/2)}{\omega^2_j}$, where $x=(x_1,\ldots,x_d)$ and $\omega=(\omega_1,\ldots,\omega_d)$.
\end{enumerate}
The following are some examples of translation invariant kernels on $\bb{R}^d$ that are not $c_0$-universal but cc-universal. These kernels satisfy $\emph{supp}(\hat{\psi})\subsetneq\bb{R}^d$ and $(\emph{supp}(\hat{\psi}))^\circ\ne\emptyset$.
\begin{enumerate}
\item [(4)] Sinc kernel, $\psi(x)=\prod^d_{j=1}\frac{\sin \sigma x_j}{x_j},\,\sigma\in\bb{R}\,:\,$ $\hat{\psi}(\omega)=\left(\frac{\pi}{2}\right)^{d/2}\prod^d_{j=1}\mathds{1}_{[-\sigma,\sigma]}(\omega_j)$ and $\emph{supp}(\hat{\psi})=[-\sigma,\sigma]^d\subsetneq\bb{R}^d$.
\item [(5)] Sinc-squared kernel, $\psi(x)=\prod^d_{j=1}\frac{\sin^2 x_j}{x^2_j}\,:\,$ $\hat{\psi}(\omega)=\frac{(2\pi)^{d/2}}{4^d}\prod^d_{j=1}(1-|\omega_j|)\mathds{1}_{[-1,1]}(\omega_j)$ and $\emph{supp}(\hat{\psi})=[-1,1]^d\subsetneq\bb{R}^d$.
\end{enumerate}
\end{example}
 The following remarks can be made about Proposition~\ref{pro:rd}.
\begin{remark}\label{rem:Rd}
(a) Theorem 6.8 in \citet{Wendland-05} states that: if $(\emph{supp}(\Lambda))^\circ\ne\emptyset$, then $k(x,y)=\psi(x-y)$ is strictly pd. By Proposition~\ref{pro:rd}(a), this means a strictly pd kernel need not be $c_0$-universal and therefore need not satisfy the condition in (\ref{Eq:strictpd}), i.e., strictly pd is not a sufficient condition for (\ref{Eq:strictpd}) to hold (see Remark~\ref{rem:strictlypd}(a)). As an example, a sinc-squared kernel is strictly pd but not $c_0$-universal (see Example~\ref{Ex:c0-universal}).
\vspace{2mm}\\
(b) In Proposition~\ref{pro:strictpd}(d), we have shown that strictly pd is a necessary condition for a kernel to be \emph{$c_0$-} or cc-universal. From the above remark, it is clear that $k$ being strictly pd does not imply it is $c_0$-universal. But does it imply $k$ is cc-universal? In general, it is not clear whether this is true. However, if  $\psi\in C_b(\bb{R}^d)\cap L^1(\bb{R}^d)$ is strictly pd, then $k(x,y)=\psi(x-y)$ is cc-universal. This follows from \citet[Theorem 6.11, Corollary 6.12]{Wendland-05}: if $\psi\in C_b(\bb{R}^d)\cap L^1\bb{R}^d)$ is strictly pd, then $0\ne\hat{\psi}\in L^1(\bb{R}^d)$, $\hat{\psi}\ge 0$ and $(\emph{supp}(\hat{\psi}))^\circ\ne\emptyset$, which by Proposition~\ref{pro:rd}(c) implies $k$ is cc-universal.\vspace{2mm}\\
(c) Is the converse to Proposition~\ref{pro:rd}(c) true? I.e., if $k$ is cc-universal, then does $(\emph{supp}(\Lambda))^\circ\ne\emptyset$ hold? Let $X=\bb{R}$. Suppose $(\emph{supp}(\Lambda))^\circ=\emptyset$, which means $\emph{supp}(\Lambda)$ is of the form $\{0,\pm\omega_1,\pm\omega_2,\ldots\}$, where $0\ne \omega_j\in\bb{R}$ for all $j$. Let us assume that there exists a non-zero entire function, $h$ on $\bb{C}$ that satisfies (i) $h(\omega_j)=0,\,\forall\,j$ and (ii) for each $N\in\bb{N}$, there is a $C_N$ such that
\begin{equation}\label{Eq:exponential-type}
|h(\zeta)|\le \frac{C_Ne^{R|\emph{Im}\,\zeta|}}{(1+|\zeta|)^N},\nonumber
\end{equation}
for all $\zeta\in\bb{C}$ and some $R>0$. Here $\emph{Im}\,\zeta$ represents the imaginary part of $\zeta$. By the Paley-Wiener theorem \cite[Theorem IX.11, p. 16]{Reed-72}, $\check{h}\in C_0(\bb{R})$ is an infinitely differentiable function on $\bb{R}$ and $\emph{supp}(\check{h})\subset\{x\in\bb{R}\,:\,|x|\le R\}$. Define $d\mu(x)=\check{h}(x)\,dx$. It is easy check that
\begin{eqnarray}
\int\!\!\int_{\bb{R}}k(x,y)\,d\mu(x)\,d\mu(y)&\!\!=\!\!&\int\!\!\int_{\bb{R}}k(x,y)\check{h}(x)\check{h}(y)\,dx\,dy
=2\pi\int_{\bb{R}}\left\vert\hat{\check{h}}(\omega)\right\vert^2\,d\Lambda(\omega)\nonumber\\
&\!\!=\!\!&2\pi\int_{\bb{R}}\left\vert h(\omega)\right\vert^2\,d\Lambda(\omega)=2\pi\sum_j |h(\omega_j)|^2\,\Lambda(\{\omega_j\})=0.\nonumber
\end{eqnarray}
This means there exists $0\ne\mu\in M_{bc}(\bb{R})$ such that $\int\!\!\int_{\bb{R}}k(x,y)\,d\mu(x)\,d\mu(y)=0$, which means $k$ is not cc-universal, by Proposition~\ref{pro:strictpd}(b).
Therefore, if $k$ is cc-universal, then $(\emph{supp}(\Lambda))^\circ\ne\emptyset$, under the assumption that there exists an $h$ that satisfies (i) and (ii) shown above. The construction of such an $h$ is not straightforward for any $k$, and therefore it is not clear whether the above converse is true in general. 
\par On the other hand, \citet[Example 5]{Sriperumbudur-09a} have shown that if $k$ is a periodic kernel (these kernels satisfy $(\emph{supp}(\Lambda))^\circ=\emptyset$), then such an $h$ defined on $\bb{R}$ can be constructed. This means if $k$ is cc-universal on $\bb{R}$, then it is not periodic on $\bb{R}$. However, this does not rule out the case of $k$ being cc-universal but aperiodic such that $(\emph{supp}(\Lambda))^\circ=\emptyset$.
\end{remark}
A summary of results, based on Proposition~\ref{pro:rd} and Remark~\ref{rem:Rd}, for the case of kernels satisfying ($A_1$), is shown in Figure~\ref{Fig:results}(c).

\subsection{Translation invariant kernels on $\bb{T}^d$: ($A_2$)}\label{subsubsec:a2}
First note that since $\bb{T}^d$ is a compact metric space, the notions of \emph{c-universality}, \emph{cc-universality} and \emph{$c_0$-universality} are equivalent. \citet[Corollary 11]{Steinwart-01} provided a sufficient condition for a Fourier kernel to be \emph{c-universal}. In Proposition~\ref{pro:td}, we show that this condition is also necessary. Using this result, we then show that the converse to Proposition~\ref{pro:strictpd}(d) is not true. Before we present the result on the characterization of \emph{c-universality} of kernels in ($A_2$), we state Bochner's theorem that characterizes pd functions, $\psi$ on $\bb{T}^d$.
\begin{theorem}[Bochner]\label{Thm:Td}
$\psi\in C(\bb{T}^d)$ is pd if and only if
\begin{equation}\label{Eq:td}
\psi(x)=\sum_{n\in\bb{Z}^d} A_\psi(n)e^{ix^Tn},\,x\in\bb{T}^d,
\end{equation}
where $A_\psi:\bb{Z}^d\rightarrow\bb{R}_+$, $A_\psi(-n)=A_\psi(n)$ and $\sum_{n\in\bb{Z}^d}A_\psi(n)<\infty$. $A_\psi$ are called the Fourier series coefficients of $\psi$.
\end{theorem}
\begin{proposition}[Translation invariant kernels on $\bb{T}^d$]\label{pro:td}
Suppose \emph{($A_2$)} holds. Then, $k$ is c-universal if and only if $A_\psi(n)>0,\,\forall\,n\in\bb{Z}^d$.
\end{proposition}
\begin{proof}
($\,\Leftarrow\,$) Consider $\int\!\!\int_{\bb{T}^d} k(x,y)\,d\mu(x)\,d\mu(y)$ for $0\ne\mu\in M_b(\bb{T}^d)$. Substituting for $k$ as in ($A_2$) and for $\psi$ as in (\ref{Eq:td}), we have
\begin{eqnarray}\label{Eq:td-suff}
B:=\int\!\!\int_{\bb{T}^d}k(x,y)\,d\mu(x)\,d\mu(y)&\!\!=\!\!&\int\!\!\int_{\bb{T}^d}\sum_{n\in\bb{Z}^d}A_\psi(n)e^{i(x-y)^Tn}\,d\mu(x)\,d\mu(y)\nonumber\\
&\!\!\stackrel{(a)}{=}\!\!& \sum_{n\in\bb{Z}^d}A_\psi(n)\int_{\bb{T}^d}e^{ix^Tn}\,d\mu(x)\int_{\bb{T}^d}e^{-iy^Tn}\,d\mu(y)\nonumber\\
&\!\!\stackrel{(b)}{=}\!\!& (2\pi)^{2d}\sum_{n\in\bb{Z}^d}A_\psi(n)\overline{A_\mu(n)}A_\mu(n)\nonumber\\
&\!\!=\!\!&(2\pi)^{2d}\sum_{n\in\bb{Z}^d}A_\psi(n)|A_\mu(n)|^2,
\end{eqnarray}
where Fubini's theorem is invoked in $(a)$ and 
\begin{equation}\label{Eq:fourier-td}
A_\mu(n):=(2\pi)^{-d}\int_{\bb{T}^d}e^{-in^Tx}\,d\mu(x),\,n\in\bb{Z}^d,
\end{equation} 
is used in $(b)$. Note that $A_\mu$ is the Fourier transform of $\mu$ in $\bb{T}^d$. Since $A_\psi(n)>0,\,\forall\,n\in\bb{Z}^d$, we have $B>0$, which by Proposition~\ref{pro:strictpd}(a) implies $k$ is \emph{c-universal}.\vspace{1mm}\\
($\,\Rightarrow\,$) Proving necessity is equivalent to proving that if $A_\psi(n)=0$ for some $n=n_0$, then there exists $0\ne \mu\in M_b(\bb{T}^d)$ such that $\int\!\!\int_{\bb{T}^d} k(x,y)\,d\mu(x)\,d\mu(y)=0$. 
\par Let $A_\psi(n)=0$ for some $n= n_0$. Define $d\mu(x)=2\alpha\cos(x^Tn_0)\,dx,\,\alpha\in\bb{R}\backslash\{0\}$. By (\ref{Eq:fourier-td}), we get $A_\mu(n)=\alpha\delta_{n_0}(n)$, where $\delta$ represents the Kronecker delta. This means $\mu\ne 0$. Using $A_\psi$ and $A_\mu$ in (\ref{Eq:td-suff}), it is easy to check that $\int\!\!\int_{\bb{T}^d} k(x,y)\,d\mu(x)\,d\mu(y)=0$. Therefore, $k$ is not \emph{c-universal}.
\end{proof}
Note that Proposition~\ref{pro:td} provides an easy to check condition for the \emph{c-universality} of translation invariant kernels on $\bb{T}^d$.
\begin{example}\label{Exm:td}
The following are some examples of translation invariant kernels on $\bb{T}$ that are c-universal (and therefore $c_0$-universal and cc-universal).
\begin{enumerate}
\item [(1)] Poisson kernel, $\psi(x)=\frac{1-\sigma^2}{\sigma^2-2\sigma\cos x+1},\,0<\sigma<1$ with $A_\psi(n)=\sigma^{|n|},\,n\in\bb{Z}$.
\item [(2)] $\psi(x)=e^{\alpha\cos x}\cos(\alpha\sin x),\,0<\alpha\le 1$ with $A_\psi(0)=1$ and $A_\psi(n)=\frac{\alpha^{|n|}}{2|n|!},\,\forall\,n\ne 0$.
\item [(3)] $\psi(x)=(\pi-(x)_{mod\,\,2\pi})^2$ with $A_\psi(0)=\frac{\pi^2}{3}$ and $A_\psi(n)=\frac{2}{n^2},\,\forall\,n\ne 0$.
\end{enumerate}
Some examples of translation invariant kernels on $\bb{T}$ that are not c-universal (and therefore not $c_0$-universal and not cc-universal) are:
\begin{enumerate}
\item[(4)] Dirichlet kernel, $\psi(x)=\frac{\sin\frac{(2l+1)x}{2}}{\sin\frac{x}{2}},\,l\in\bb{N}$ with $A_\psi(n)=1$ for $n\in\{0,\pm 1,\ldots,\pm l\}=:D$ and $A_\psi(n)=0$ for $n\notin D$.
\item[(5)] Fej\'{e}r kernel, $\psi(x)=\frac{1}{l+1}\frac{\sin^2\frac{(l+1)x}{2}}{\sin^2\frac{x}{2}},\,l\in\bb{N}$ with $A_\psi(n)=1-\frac{|n|}{l+1}$ for $n\in D$ and $A_\psi(n)=0$ for $n\notin D$.
\end{enumerate}
\end{example}
\textbf{$c$-universal kernels vs. Strictly pd kernels:} We have shown in Proposition~\ref{pro:strictpd}(d) that strictly pd is a necessary condition for $k$ to be \emph{c-}, \emph{cc-} or \emph{$c_0$-universal}. However, the converse is not true (see Remark~\ref{rem:strictlypd}(a)), which is based on Proposition~\ref{pro:td} and the following result in Theorem~\ref{Thm:spd-sphere}. Before we state the result, we need some definitions.
\par For natural numbers $m$ and $n$ and a set $A$ of integers, $m+nA:=\{j\in\bb{Z}\,|\,j=m+na,\,a\in A\}$. An increasing sequence $\{c_l\}$ of nonnegative integers is said to be \emph{prime} if it is not contained in any set of the form $p_1\bb{N}\cup p_2\bb{N}\cup\cdots\cup p_n\bb{N}$, where $p_1,p_2,\ldots,p_n$ are prime numbers. Any infinite increasing sequence of prime numbers is a trivial example of a prime sequence. We write $\bb{N}^0_n:=\{0,1,\ldots,n\}$.
\begin{theorem}[\citet{Menegatto-95}]\label{Thm:spd-sphere}
Let $\psi$ be a pd function on $\bb{T}$ of the form in (\ref{Eq:td}). Let $\overline{N}:=\{|n|:A_\psi(n)>0,\,n\in\bb{Z}\}\subset\bb{N}\cup\{0\}$. Then $\psi$ is strictly pd if $\overline{N}$ has a subset of the form $\cup^\infty_{l=0}(b_l+c_l\bb{N}^0_l)$, in which $\{b_l\}\cup\{c_l\}\subset\bb{N}$ and $\{c_l\}$ is a prime sequence.
\end{theorem}
Suppose $\psi$ be such that $\overline{N}\subsetneq\bb{N}\cup\{0\}$ has a subset of the form as mentioned in Theorem~\ref{Thm:spd-sphere}. Clearly, $\psi$ is strictly pd. However, it is not \emph{c-universal} as Proposition~\ref{pro:td} 
states that $k$ is \emph{c-universal} if and only if $\overline{N}=\bb{N}\cup\{0\}$.
\par A summary of results for kernels of the type ($A_2$) is shown in Figure~\ref{Fig:results}(b).

\subsection{Radial kernels on $\bb{R}^d$: ($A_3$)}\label{subsubsec:a3}
The following result provides an easily checkable characterization for $k$ to be \emph{$c_0$}- and \emph{cc-universal} (\emph{c-universality} is not considered as $X=\bb{R}^d$ is not compact) when $k$ satisfies ($A_3$).
\begin{proposition}[Radial kernels on $\bb{R}^d$]\label{pro:a3}
Suppose \emph{($A_3$)} holds. Then the following conditions are equivalent.
\begin{itemize}
\item[(a)] $k$ is $c_0$-universal.
\item[(b)] $\emph{supp}(\nu)\ne\{0\}$.
\item[(c)] $k$ is strictly pd.
\item[(d)] $k$ is cc-universal.
\end{itemize}
\end{proposition}
\begin{proof}
$(a)\Rightarrow (d)$ by Remark~\ref{rem:universal}(c), $(d)\Rightarrow (c)$ by Proposition~\ref{pro:strictpd}(d) and $(c)\Leftrightarrow (b)$ by  \citet[Theorem 7.14]{Wendland-05}. Now, we show $(b)\Rightarrow (a)$. 
\par Consider $\int\!\!\int_{\bb{R}^d} k(x,y)\,d\mu(x)\,d\mu(y)$ with $k$ as in (\ref{Eq:schoenberg}), given by
\begin{eqnarray}\label{Eq:radial}
B:=\int\!\!\int_{\bb{R}^d} k(x,y)\,d\mu(x)\,d\mu(y)\!\!&=&\!\!\int\!\!\int_{\bb{R}^d}\int^\infty_{0} e^{-t\Vert x-y\Vert^2_2}\,d\nu(t)\,d\mu(x)\,d\mu(y)\nonumber\\
\!\!&\stackrel{(e)}{=}&\!\!\int^\infty_{0}\left[\int\!\!\int_{\bb{R}^d}e^{-t\Vert x-y\Vert^2_2}\,d\mu(x)\,d\mu(y)\right]\,d\nu(t)\nonumber\\
\!\!&\stackrel{(f)}{=}&\!\!\int^\infty_{0}\frac{1}{(2t)^{d/2}}\left[\int_{\bb{R}^d}|\hat{\mu}(\omega)|^2e^{-\frac{\Vert\omega\Vert^2_2}{4t}}\,d\omega\right]\,d\nu(t)\nonumber\\
\!\!&\stackrel{(g)}{=}&\!\!\int_{\bb{R}^d}|\hat{\mu}(\omega)|^2\left[\int^\infty_{0}\frac{1}{(2t)^{d/2}}e^{-\frac{\Vert\omega\Vert^2_2}{4t}}\,d\nu(t)\right]\,d\omega,
\end{eqnarray}
where Fubini's theorem is invoked in $(e)$ and $(g)$, while (\ref{Eq:translation}) is invoked in $(f)$. Since $\text{supp}(\nu)\ne\{0\}$, the inner integral in (\ref{Eq:radial}) is positive for every $\omega\in\bb{R}^d$ and so $B>0$. Therefore $k$ is \emph{$c_0$-universal} by Proposition~\ref{pro:strictpd}.
\end{proof}
The above result shows that the notions of \emph{$c_0$-universality}, \emph{cc-universality} and strict positive definiteness are equivalent for the class of radial kernels on $\bb{R}^d$.
\begin{example}\label{Exm:radial}
The following radial kernels on $\bb{R}^d$ have $\emph{supp}(\nu)\ne\{0\}$ and therefore are $c_0$-universal, cc-universal and strictly pd.
\begin{enumerate}
\item[(1)] Gaussian, $k(x,y)=e^{-\sigma \Vert x-y\Vert^2_2},\,\sigma>0$. Note that $\nu=\delta_\sigma$ in (\ref{Eq:schoenberg}), where $\delta_\sigma$ represents a Dirac measure at $\sigma$. Clearly $\emph{supp}(\nu)=\{\sigma\}\ne\{0\}$.
\item[(2)] Inverse multiquadratic, $k(x,y)=(c^2+\Vert x-y\Vert^2_2)^{-\beta},\,\beta>0,\,c>0$, obtained by choosing $d\nu(t)=\frac{1}{\Gamma(\beta)}t^{\beta-1}e^{-c^2t}\,dt$ in (\ref{Eq:schoenberg}). It is easy to verify that $\emph{supp}(\nu)\ne\{0\}$. 
\end{enumerate}
\end{example}
A summary of results for kernels of the type ($A_3$) is shown in Figure~\ref{Fig:results}(d).

\subsection{Kernels of type ($A_4$)}\label{subsubsec:a4}
We now consider the characterization of \emph{c-}, \emph{cc-} and \emph{$c_0$-universality} for ($A_4$). 
\begin{proposition}[Kernels of type ($A_4$)]\label{pro:a4}
Suppose \emph{($A_4$)} holds.
\begin{itemize}
\item[(a)] $k$ is c-universal (\emph{resp.} cc-universal) if and only if for any $0\ne\mu\in M_{b}(X)$ (\emph{resp.} $0\ne\mu\in M_{bc}(X)$), there exists some $j\in I$ for which $\int_X \phi_j\,d\mu\ne 0$.
\item[(b)] Let $k(\cdot,x)\in C_0(X),\,\forall\,x\in X$. Then $k$ is $c_0$-universal if and only if for any $0\ne\mu\in M_b(X)$, there exists some $j\in I$ for which $\int_X \phi_j\,d\mu\ne 0$.
\end{itemize}
\end{proposition}
\begin{proof}
We first prove $(b)$. The proof for \emph{c-universality} in $(a)$ is trivial as it follows from $(b)$, while the proof for \emph{cc-universality} in $(a)$ is exactly the same as that of $(b)$ with $M_b(X)$ replaced by $M_{bc}(X)$. Let us consider
\begin{equation}\label{Eq:a4}
\int\!\!\int_X k(x,y)\,d\mu(x)\,d\mu(y)=\int\!\!\int_X \sum_{j\in I}\phi_j(x)\phi_j(y)\,d\mu(x)\,d\mu(y)\stackrel{(c)}{=}\sum_{j\in I}\left\vert\int_X\phi_j(x)\,d\mu(x)\right\vert^2,
\end{equation}
where we have invoked Fubini's theorem in $(c)$.\vspace{1mm}\\
$(b)$ ($\,\Leftarrow\,$) Suppose for any $0\ne\mu\in M_b(X)$, there exists some $j\in I$ for which $\int_X\phi_j\,d\mu\ne 0$. Then, from (\ref{Eq:a4}), it is clear that $\int\!\!\int_X k(x,y)\,d\mu(x)\,d\mu(y)>0,\,\forall\,0\ne\mu\in M_b(X)$ and therefore $k$ is \emph{$c_0$-universal}, which follows from Proposition~\ref{pro:strictpd}(c).
\vspace{1mm}\\
($\,\Rightarrow\,$) Suppose there exists a non-zero measure, $\mu\in M_b(X)$ for which $\int_X \phi_j\,d\mu=0$ for any $j\in I$. By (\ref{Eq:a4}), this means there exists a $0\ne\mu\in M_b(X)$ for which $\int\!\!\int_X k(x,y)\,d\mu(x)\,d\mu(y)=0$, i.e., $k$ is not \emph{$c_0$-universal} (by Proposition~\ref{pro:strictpd}(c)).
\end{proof}
The conditions in Proposition~\ref{pro:a4} are not always easy to check. However, for the case of \emph{Taylor kernels} \citep[Lemma 4.8]{Steinwart-08}, which include the exponential kernel, simple, easy to check sufficient conditions can be obtained as shown in Corollary~\ref{cor:taylor}. Although this result is exactly the same as Corollary 4.57 in \citet{Steinwart-08}, we present a different proof (we would like to remind the reader that our characterization of \emph{c-universality} is different from the one provided by \citet{Steinwart-01} and therefore the proof is different; see Remark~\ref{rem:universal}(a)).
\begin{corollary}[Universal Taylor kernels]\label{cor:taylor}
Let $X:=\{x\in\bb{R}^d:\Vert x\Vert_2<\sqrt{r}\}$, where $r\in (0,\infty]$. Let $f(t)=\sum^\infty_{n=0}a_nt^n,\,t\in(-r,r)$. If $a_n>0,\,\forall\,n\ge 0$, then $k(x,y)=f(x^Ty),\,x,y\in X$, is $c$-universal on every compact subset of $X$.
\end{corollary}
\begin{proof}
From the proof of Lemma 4.8 in \citet{Steinwart-08}, we have
\begin{equation}\label{Eq:dot}
k(x,y)=f(x^Ty)=\sum^\infty_{n=0}a_n\left(x^Ty\right)^n=\sum_{\alpha\in \bb{N}^d} a_{|\alpha|}c_\alpha x^\alpha y^\alpha,
\end{equation}
where $\alpha:=(\alpha_j:j\in\bb{N}_d)$, $|\alpha|:=\sum_{j\in \bb{N}_d} \alpha_j$, $c_\alpha:=\frac{n!}{\prod^d_{j=1}\alpha_j!}$, $x=(x_1,\ldots,x_d)$ and $x^\alpha:=\prod^d_{j=1}(x_j)^{\alpha_j}$. From (\ref{Eq:dot}), it is clear that $k(x,y)=\sum_{\alpha\in \bb{N}^d}\phi_\alpha(x)\phi_\alpha(y),\,x,y\in X$, where $\phi_\alpha(x)=\sqrt{a_{|\alpha|}c_\alpha}x^\alpha$.
Let $a_{|\alpha|}>0$ for all $\alpha\in\bb{N}^d$. Then it is clear that for any $0\ne\mu\in M_b(X)$, there exists $\alpha\in \bb{N}^d$ such that $\int_X x^\alpha\,d\mu(x)\ne 0$. Therefore, by Proposition~\ref{pro:a4}, $k$ is \emph{c-universal}.
\end{proof}
Examples of kernels that satisfy the conditions in Corollary~\ref{cor:taylor} and therefore are \emph{c-universal} include the exponential kernel, $k(x,y)=\exp(x^Ty),\,x,y\in\bb{R}^d$, binomial kernel, $k(x,y)=(1-x^Ty)^{-\beta},\,\beta>0$, defined on $X\times X$, where $X:=\{x\in\bb{R}^d\,:\,\Vert x\Vert_2<1\}$, etc. See Examples 4.9 and 4.11 in \citet{Steinwart-08}).\vspace{2.5mm}
\par\noindent To summarize, in this section, by showing the relation between various notions of universality and the injective RKHS embedding of finite signed Radon measures, we have presented a novel measure embedding point of view of universality compared to its well-known function approximation view point. 
Since the RKHS embedding of finite signed Radon measures generalizes the concept of RKHS embedding of Radon probability measures, the latter being related to \emph{characteristic kernels} \citep{Fukumizu-04,Fukumizu-08a,Sriperumbudur-08}, in the following section, we relate the notion of universality to characteristic kernels.

\section{Characteristic Kernels and Universality}\label{Sec:characteristic}
Recent studies in machine learning have considered the mapping of random variables into a suitable RKHS and showed that this provides a powerful and straightforward method of dealing with higher-order statistics of the variables. Using their RKHS mappings, for sufficiently \emph{rich} RKHSs, it becomes possible to test for homogeneity \citep{Gretton-06}, independence \citep{Gretton-08}, conditional independence \citep{Fukumizu-08a}, to find the most predictive subspace in regression~\citep{Fukumizu-04}, etc. Key to the above applications is the notion of a \emph{characteristic kernel}~---~
defined below~---~which gives rise to an RKHS that is sufficiently rich in the sense required above.
\begin{definition}[Characteristic kernel]\label{def:characteristic}
Let $X$ be a topological space, $\bb{P}$ be a Borel probability measure on $X$ and $k$ be a measurable, bounded kernel on $X$. Then $k$ is said to be characteristic if the embedding,
\begin{equation}\label{Eq:charac}
\vspace{-1mm}
\bb{P}\mapsto\int_X k(\cdot,x)\,d\bb{P}(x),
\vspace{-1mm}
\end{equation}
is injective.
\end{definition}
Since the embedding in (\ref{Eq:charac}) is a special case of the embedding in (\ref{Eq:signedmeasure}), and the injectivity of the embedding in (\ref{Eq:signedmeasure}) is related to universality (see Section~\ref{Sec:universality}), we now relate universal and characteristic kernels. 
\subsection{Main results}\label{subsec:main-char}
\citet{Gretton-06} have shown that a \emph{c-universal} kernel is \emph{characteristic}. Besides this result, not much is known or understood about the relation between characteristic and universal kernels. The following result not only provides the same result obtained by \citet{Gretton-06}, but also generalizes it for non-compact $X$.
\begin{proposition}[Universal and characteristic kernels$-$I]\label{pro:relation-1}
Suppose the assumptions in Theorem~\ref{Thm:c0-universal} hold. If $k$ is $c$-, $cc$- or $c_0$-universal, then it is characteristic to the set of probability measures contained in $M_b(X)$, $M_{bc}(X)$ or $M_b(X)$, respectively.\vspace{-.5mm}
\end{proposition}
\begin{proof}
The proof is trivial and follows from Theorem~\ref{Thm:c0-universal} and Definition~\ref{def:characteristic}.
\end{proof}
Now, one can ask when the converse to Proposition~\ref{pro:relation-1} is true. The following result answers this question for some special classes of kernels.\vspace{-.5mm}
\begin{proposition}[Universal and characteristic kernels$-$II]\label{pro:relation-2}
The following hold:\vspace{-1mm}
\begin{itemize}
\item[(a)] Suppose \emph{($A_1$)} holds with $\psi\in C_0(\bb{R}^d)$. Then, $k$ is $c_0$-universal if and only if it is characteristic to the set of all Borel probability measures on $\bb{R}^d$.
\item[(b)] Suppose \emph{($A_2$)} holds. Then, $k$ is c-universal if it is characteristic to the set of all Borel probability measures on $\bb{T}^d$ and $A_\psi(0)>0$, where $A_\psi$ is defined in (\ref{Eq:td}).
\item[(c)] Suppose \emph{($A_3$)} holds. Then, $k$ is cc-universal if and only if it is characteristic to the set of all Borel probability measures on $\bb{R}^d$.\vspace{-.5mm}
\end{itemize}
\end{proposition}
\begin{proof}$(a)$ Suppose $k$ is \emph{$c_0$-universal}. Then, by Proposition~\ref{pro:relation-1}, $k$ is characteristic to $M^1_+(\bb{R}^d)$. Conversely, if $k$ is characteristic to $M^1_+(\bb{R}^d)$, we have $\text{supp}(\Lambda)=\bb{R}^d$ which follows from Theorem 7 in \citet{Sriperumbudur-08}. 
The result therefore follows from Proposition~\ref{pro:rd}(a).\vspace{1mm}\\
$(b)$ \citet[Theorem 8]{Fukumizu-08b} and \citet[Theorem 14]{Sriperumbudur-09a} have shown that $k$ is characteristic to $M^1_+(\bb{T}^d)$ if and only if $A_\psi(0)\ge 0$, $A_\psi(n)>0,\,\forall\,n\in\bb{Z}^d\backslash\{0\}$. Therefore, if $k$ is characteristic with $A_\psi(0)>0$, then it is \emph{c-universal} by Proposition~\ref{pro:td}.\vspace{1mm}\\
$(c)$ If $k$ is \emph{cc-universal}, then by Proposition~\ref{pro:a3}, it is \emph{$c_0$-universal}, and thus characteristic to $M^1_+(\bb{R}^d)$ by Proposition~\ref{pro:relation-1}. To prove the converse, we need to prove that if $k$ is not \emph{cc-universal}, then it is not characteristic to $M^1_+(\bb{R}^d)$. If $k$ is not \emph{cc-universal}, then by Proposition~\ref{pro:a3}, we have $\text{supp}(\nu)=\{0\}$ (see (\ref{Eq:schoenberg}) for the definition of $\nu$), which means the kernel is a constant function on $\bb{R}^d\times\bb{R}^d$ and therefore not characteristic to $M^1_+(\bb{R}^d)$.\vspace{-7mm}
\end{proof}
\begin{remark}\label{rem:charac}
(a) If $k$ is $c_0$-universal, then $k$ is characteristic, which follows from Proposition~\ref{pro:relation-1}. In general, the converse is not true, which follows from Proposition~\ref{pro:td} and Proposition~\ref{pro:relation-2}(b). However, on the class of translation invariant kernels and radial kernels defined over $\bb{R}^d$, the converse is true, which is shown in Proposition~\ref{pro:relation-2}(a,c).\vspace{2mm}\\
(b) Although an RKHS, $\eu{H}$ can be characteristic without containing constant functions \citep[Lemma 1]{Fukumizu-08b}, Proposition~\ref{pro:relation-2}(b) shows that if $\eu{H}$ does contain constant functions (i.e., $A_\psi(0)>0$), then the class of characteristic kernels on $\bb{T}^d$ is equivalent to the class of c-universal (and, therefore, cc- and $c_0$-universal) kernels. Based on \citet[Lemma 1]{Fukumizu-08b} and \citet[Theorem 1]{Carmeli-09}, this result can be generalized to any LCH space, $X$, which says that if constant functions are included in $\eu{H}$, then characteristic kernels are equivalent to $c_0$-universal kernels.\vspace{-1mm}
\end{remark}
\par A summary of the relation between characteristic and universal kernels is shown in Figure~\ref{Fig:results}.\vspace{2mm}\\
\textbf{Characteristic kernels vs. Strictly pd kernels:} In Section~\ref{Sec:universality}, we have shown the relation between universal kernels and strictly pd kernels, while in Propositions~\ref{pro:relation-1} and \ref{pro:relation-2}, we have related universal and characteristic kernels. We now investigate the relation between characteristic and strictly pd kernels.
\par Based on Propositions~\ref{pro:rd}, \ref{pro:a3} and \ref{pro:relation-2}, it is clear that a characteristic kernel that is translation invariant or radial on $\bb{R}^d$ is strictly pd. While the converse holds for radial kernels on $\bb{R}^d$, it does not hold for translation invariant kernels on $\bb{R}^d$, which follows from Proposition~\ref{pro:relation-2} and Remark~\ref{rem:Rd}(a). Similarly, in the case of translation invariant kernels on $\bb{T}$, if a kernel is characteristic, then it is strictly pd, which follows from Theorem~\ref{Thm:spd-sphere} and Proposition~\ref{pro:relation-2}, while the converse is not true. So far, we have presented the relation between characteristic and strictly pd kernels for specific cases of kernels satisfying ($A_1$)--($A_3$), which is summarized in Figure~\ref{Fig:results}. For the general case, it is not clear whether strict pd is a necessary condition for $k$ to be characteristic. However, the following result shows that \emph{conditionally strictly pd} is a necessary condition for $k$ to be characteristic.\vspace{-.5mm}
\begin{proposition}\label{pro:cpd}
If $k$ is characteristic, then it is conditionally strictly pd.\vspace{-.5mm}
\end{proposition}
\begin{proof}
Suppose $k$ is not conditionally strictly pd. This means for some $n\ge 2$ and for mutually distinct $x_1,\ldots,x_n\in X$, there exists $\{\alpha_j\}\ne 0$ with $\sum^n_{j=1}\alpha_j=0$ such that $\sum^n_{l,j=1}\alpha_l\alpha_jk(x_l,x_j)=0$. Define $\mu:=\sum^n_{j=1}\alpha_j\delta_{x_j}$, where $\delta_{x}$ represents the Dirac measure at $x$. Clearly, $\mu$ is a finite non-zero Borel measure that satisfies $(i)$ $\int\!\!\int_X k(x,y)\,d\mu(x)\,d\mu(y)=0$ and $(ii)$ $\mu(X)=0$. Since $\mu$ is a finite non-zero Borel measure, by the Jordan decomposition theorem \citep[Theorem 5.6.1]{Dudley-02}, there exist unique positive measures $\mu^+$ and $\mu^-$ such that $\mu=\mu^+-\mu^-$ and $\mu^+\perp \mu^-$ ($\mu^+$ and $\mu^-$ are singular). By \emph{(ii)}, we have $\mu^+(X)=\mu^-(X)=:\alpha$. Define $\bb{P}=\alpha^{-1}\mu^+$ and $\bb{Q}=\alpha^{-1}\mu^-$. Clearly, $\bb{P}$ and $\bb{Q}$ are distinct Borel probability measures defined on $X$. Then, we have
\begin{eqnarray}
\left\Vert\int_X k(\cdot,x)\,d\bb{P}(x)-\int_X k(\cdot,x)\,d\bb{Q}(x)\right\Vert^2_\eu{H}\!\!\!&\stackrel{(a)}{=}&\!\!\!\int\!\!\!\int_{X}k(x,y)\,d(\bb{P}-\bb{Q})(x)\,d(\bb{P}-\bb{Q})(y)
\nonumber\\
\!\!\!&=&\!\!\!\alpha^{-2}\int\!\!\!\int_{M}k(x,y)\,d\mu(x)\,d\mu(y)\stackrel{(b)}{=}0,\nonumber
\end{eqnarray}
where Lemma~\ref{lem:interchange} is invoked in \emph{(a)} and \emph{(b)} is obtained by invoking \emph{(i)}. So, there exist $\bb{P}\ne\bb{Q}$ such that $\int_X k(\cdot,x)\,d\bb{(P-Q)}(x)=0$, i.e., $k$ is not characteristic.
\vspace{-.5mm}
\end{proof}
The converse to Proposition~\ref{pro:cpd} is however not true. 
\par So far, we presented the relation between characteristic kernels and universal kernels and showed that for any LCH space, $X$, the characteristic property is a weaker notion than \emph{$c_0$-universality}. Although such a weaker notion is sufficient to make the embedding in (\ref{Eq:charac}) injective, in the following section, we show that the stronger notion of \emph{$c_0$-universality} is required to study an important property of the ``probability metric" associated with the embedding in (\ref{Eq:charac}). 
\subsection{Metrization of weak topology on $M^1_+(X)$}\label{subsec:metric}
Let $X$ be a Polish space.\footnote{A topological space $(X,\tau)$ is called a Polish space if the topology $\tau$ has a countable basis and there exists a complete metric defining $\tau$.} Based on the embedding, $\bb{P}\mapsto \int_X k(\cdot,x)\,d\bb{P}(x),\,\bb{P}\in M^1_+(X)$, \citet{Gretton-06} proposed the following pseudometric,
\begin{equation}\label{Eq:mmd}
\gamma_k(\bb{P},\bb{Q}):=\left\Vert\int_X k(\cdot,x)\,d\bb{P}(x)-\int_X k(\cdot,x)\,d\bb{Q}(x)\right\Vert_\eu{H},
\end{equation}
on $M^1_+(X)$, called the \emph{maximum mean discrepancy} (MMD). Note that when $k$ is \emph{characteristic}, $\gamma_k$ is a metric on $M^1_+(X)$. One immediate question that naturally arises is ``how is MMD related to other metrics on $M^1_+(X)$, such as the Prohorov metric, Dudley metric, Wasserstein-Kantorovich metric, total variation metric, etc?" This is a question of both theoretical and practical importance.
\par For example, let us consider the problem of estimating an unknown density based on finite random samples drawn i.i.d. from it. The quality of the estimate is measured by determining the distance between the estimated density and the true density. Given two probability metrics, $\rho_1$ and $\rho_2$, one might want to use the \emph{stronger}\footnote{Two metrics $\rho_1:Y\times Y\rightarrow\bb{R}_+$ and $\rho_2:Y\times Y\rightarrow \bb{R}_+$ are said to be equivalent if $\rho_1(x,y)=0\Leftrightarrow\rho_2(x,y)=0,\,\forall\,x,y\in Y$. On the other hand, $\rho_1$ is said to be stronger than $\rho_2$ if $\rho_1(x,y)=0\Rightarrow\rho_2(x,y)=0,\,\forall\,x,y\in Y$ but not vice-versa. If $\rho_1$ is stronger than $\rho_2$, then we say $\rho_2$ is weaker than $\rho_1$.} of the two to determine this distance, as the convergence of the estimated density to the true density in the stronger metric implies the convergence in the weaker metric, while the converse is not true. On the other hand, one might need to use a metric of weaker topology (i.e., coarser topology) to show convergence of some estimators, as the convergence might not occur w.r.t.~a metric of strong topology. This motivates a deeper analysis of the relation between probability metrics, e.g., as mentioned before, the relation between MMD and other popular probability metrics to, e.g., determine which metrics are stronger respectively weaker.
\par Recently, \citet{Sriperumbudur-09a} studied the relation between MMD and other probability metrics such as the Prohorov distance, Dudley metric, Wasserstein distance and total variation distance and showed that MMD is weaker than all these other metrics. This means that the topology induced by MMD is coarser than the topology induced by all these other metrics on $M^1_+(X)$. It is well known that the Prohorov and Dudley metrics induce a topology that coincides with the \emph{weak topology} (also called the \emph{weak}-$^\ast$ (weak-star) topology) on $M^1_+(X)$, defined as the weakest topology such that the map $\bb{P}\mapsto\int_X f\,d\bb{P}$ is continuous for all $f\in C_b(X)$. This naturally leads to the question, ``For what $k$ does the topology induced by MMD coincide with the weak topology?" In other words, ``For what $k$ is MMD equivalent to the Prohorov and Dudley metrics?" Although we arrived at this question motivated by an application, this question on its own is theoretically interesting and important in probability theory, especially in proving central limit theorems. Before we answer it (this question was answered for compact Hausdorff, $X$ and $X=\bb{R}^d$ in \citet[Section 5]{Sriperumbudur-09a}, whereas in the following, we answer it for general LCH spaces), we need some preliminaries.
\par The \emph{weak topology} on $M^1_+(X)$ is the weakest topology such that the map $\bb{P}\mapsto\int_X f\,d\bb{P}$ is continuous for all $f\in C_b(X)$. A sequence of measures is said to \emph{converge weakly} to $\bb{P}$, written as $\bb{P}_n\stackrel{w}{\rightarrow}\bb{P}$, if and only if $\int_X f\,d\bb{P}_n\rightarrow \int_X f\,d\bb{P}$ for every $f\in C_b(X)$. A metric $\gamma$ on $M^1_+(X)$ is said to \emph{metrize} the weak topology if the topology induced by $\gamma$ coincides with the weak topology, which is defined as follows: if, for $\bb{P},\bb{P}_1,\bb{P}_2,\ldots\in M^1_+(X)$, ($\bb{P}_n\stackrel{w}{\rightarrow}\bb{P}\Leftrightarrow\gamma(\bb{P}_n,\bb{P})\stackrel{n\rightarrow\infty}{\longrightarrow} 0$) holds, then the topology induced by $\gamma$ coincides with the weak topology.
\begin{proposition}\label{pro:weak}
Let $X$ be an LCH space and $k$ be $c_0$-universal. Then, the topology induced by $\gamma_k$ coincides with the weak topology on $M^1_+(X)$.
\end{proposition}
\begin{proof}
We need to show that for measures $\bb{P},\bb{P}_1,\bb{P}_2,\ldots\in M^1_+(X)$, $\bb{P}_n\stackrel{w}{\rightarrow}\bb{P}$ if and only if $\gamma_k(\bb{P}_n,\bb{P})\rightarrow 0$ as $n\rightarrow \infty$. To prove the result, we use an equivalent representation of $\gamma_k$ given by \citet[Theorem 3]{Sriperumbudur-08},
\begin{equation}\label{Eq:mmd-1}
\gamma_k(\bb{P},\bb{Q})=\sup_{\Vert f\Vert_{\eu{H}}\le 1}\left|\int_X f\,d\bb{P}-\int_X f\,d\bb{Q}\right|=\sup_{f\in\eu{H}}\frac{\left|\int_X f\,d\bb{P}-\int_X f\,d\bb{Q}\right|}{\Vert f\Vert_\eu{H}}.
\end{equation}
($\,\Leftarrow\,$) Define $\bb{P}f:=\int_X f\,d\bb{P}$. Since $k$ is \emph{$c_0$-universal}, $\eu{H}$ is dense in $C_0(X)$ w.r.t.~$\Vert\cdot\Vert_u$, i.e., for any $f\in C_0(X)$ and every $\epsilon>0$, there exists a $g\in \eu{H}$ such that $\Vert f-g\Vert_u \le \epsilon$. Therefore,
\begin{eqnarray}
\vspace{-2mm}
|\bb{P}_nf-\bb{P}f|&\!\!\!=\!\!\!&|\bb{P}_n(f-g)+\bb{P}(g-f)+(\bb{P}_ng-\bb{P}g)|\nonumber\\
&\!\!\!\le\!\!\!&  \bb{P}_n|f-g|+\bb{P}|f-g|+|\bb{P}_ng-\bb{P}g|\nonumber\\
&\!\!\!\le\!\!\!& 2\epsilon+|\bb{P}_ng-\bb{P}g|
\le 2\epsilon+\Vert g\Vert_\eu{H}\gamma_k(\bb{P}_n,\bb{P}).\vspace{-2mm}
\end{eqnarray}
Since $\gamma_k(\bb{P}_n,\bb{P})\rightarrow 0$ as $n\rightarrow\infty$ and $\epsilon$ is arbitrary, $|\bb{P}_nf-\bb{P}f|\rightarrow 0$ for any $f\in C_0(X)$. The result follows from \citet[Corollary 4.3]{Berg-84}, which says that if $\bb{P}_nf\rightarrow \bb{P}f,\,\forall\,f\in C_0(X)$, then $\bb{P}_nf\rightarrow\bb{P}f,\,\forall\,f\in C_b(X)$, i.e., $\bb{P}_n\stackrel{w}{\rightarrow}\bb{P}$.\vspace{1mm}\\
($\,\Rightarrow\,$) Suppose $\bb{P}_n\stackrel{w}{\rightarrow} \bb{P}$, i.e., $\bb{P}_nf\rightarrow\bb{P}f,\,\forall\,f\in C_b(X)$. This implies $\bb{P}_nf\rightarrow\bb{P}f,\,\forall\,f\in \eu{H}$ and therefore $\gamma_k(\bb{P}_n,\bb{P})\rightarrow 0$ as $n\rightarrow\infty$.
\end{proof}
Proposition~\ref{pro:weak} shows that if $k$ is \emph{$c_0$-universal}, then MMD induces the same topology as induced by the Prohorov and Dudley metrics and therefore is equivalent to both these metrics. This means that, although $k$ being characteristic is sufficient to guarantee $\gamma_k$ being a metric, a stronger condition on $k$, i.e., $k$ being \emph{$c_0$-universal} is required for $\gamma_k$ to metrize the weak topology on $M^1_+(X)$.
\par The following result in \citet[Theorem 23]{Sriperumbudur-09a} can be obtained as a simple corollary to Proposition~\ref{pro:weak}, wherein the question of metrization of weak topology by $\gamma_k$ is addressed only for compact Hausdorff $X$. The general non-compact case was left as an open problem, which we addressed in Proposition~\ref{pro:weak}.
\begin{corollary}[\citet{Sriperumbudur-09a}]\label{cor:weak}
Suppose $X$ is compact Hausdorff and $k$ is c-universal. Then, $\gamma_k$ metrizes the weak topology on $M^1_+(X)$.
\end{corollary}
\begin{proof}
When $X$ is compact, \emph{c-universality} and \emph{$c_0$-universality} are equivalent (see Remark~\ref{rem:universal}(c)). Therefore, the result follows from Proposition~\ref{pro:weak}.
\end{proof}
\indent To summarize, in this section, we have related the notions of \emph{universality} and \emph{characteristic} kernels by exploiting the relation between \emph{universality} and the RKHS embedding of Radon measures, which is discussed in Section~\ref{Sec:universality}. We showed that \emph{universal} and \emph{characteristic} kernels are equivalent on the class of translation invariant and radial kernels on $\bb{R}^d$. In addition, one of the open questions in \citet[Section 5]{Sriperumbudur-09a} is addressed by determining the conditions on $k$ so that $\gamma_k$ metrizes the weak topology on the space of probability measures, defined on a general non-compact $X$.
\section{Conclusions \& Discussion}\label{Sec:conclusion}
In this work, we have considered the problem of embedding finite signed Borel measures into an RKHS~---~which is a generalization of the recently studied concept of embedding Borel probability measures into an RKHS~---~and studied the conditions on the kernel under which this embedding is injective. We showed that the injectivity of this embedding is related to the notion of \emph{universality}: the embedding is injective if and only if the kernel is \emph{universal}. In other words, compared to earlier characterizations of universality \citep{Steinwart-01,Micchelli-06,Carmeli-09}, we have provided a novel characterization for \emph{universal kernels}, which is based on the measure embedding view point as opposed to the point of view of function approximation. In addition, because of this relation between \emph{universality} and the injective embedding of finite signed Borel measures, we established the relation between \emph{universal} and \emph{characteristic} kernels, the latter being related to the injective embedding of Borel probability measures into an RKHS. As an example, we showed the \emph{universal} and \emph{characteristic} property to be equivalent in the case of translation invariant and radial kernels on $\bb{R}^d$. 
\par The 
discussion in this paper has been related to the characterization of various notions of \emph{universality} wherein the RKHS, $\eu{H}$ is dense in some subset of $C(X)$ (the space of real-valued continuous functions on $X$) w.r.t.~the uniform norm (here, $X$ is a some arbitrary topological space). This means any target function, $f^\star$ in the appropriate subset of $C(X)$ can be approximated arbitrarily well by some $g\in \eu{H}$ w.r.t.~the uniform norm. There is a notion of universality, which we have not considered, called \emph{$L_p$-universality} \cite[Chapter 5]{Steinwart-08}: a measurable and bounded kernel, $k$ defined on a Hausdorff space, $X$, is said to be \emph{$L_p$-universal} if the RKHS, $\eu{H}$ induced by $k$ is dense in $L^p(X,\mu)$ w.r.t.~the \emph{$p$-norm}, defined as $\Vert f\Vert_p:=(\int_X |f(x)|^p\,d\mu(x))^{1/p}$, for all $\mu\in M^1_+(X)$ and some $p\in [1,\infty)$. Here $L^p(X,\mu)$ is the Banach space of $p$-integrable $\mu$-measurable functions on $X$. This notion of universality is more applicable in learning theory, where the target function, $f^\star$ is usually assumed to lie in $L^p(X,\mu)$ for some $p\in [1,\infty)$ and for some Borel probability measure, $\mu$. By considering this notion of universality, any $f^\star\in L^p(X,\mu)$ can be approximated arbitrarily well by some $g\in\eu{H}$ w.r.t.~the \emph{p-norm} for all Borel probability measures $\mu$ and some $p\in [1,\infty)$. In particular, \citet[Theorems 5.31, 5.36 and Corollary 5.37]{Steinwart-08} have shown that \emph{$L_p$-universality} is necessary and sufficient to achieve consistency in kernel-based learning algorithms. In this paper, we did not consider this notion of universality because unlike the other notions of universality, it is not straightforward to relate \emph{$L_p$-universality} and the RKHS embedding of measures by using the Hahn-Banach theorem (see Theorem~\ref{Thm:Hahn-Banach}). However, 
recently, \citet[Theorem 1]{Carmeli-09} have shown that $k$ is \emph{$L_p$-universal} if and only if it is \emph{$c_0$-universal}, which therefore establishes the relation between \emph{$L_p$-universality} and the RKHS embedding of measures. Using this result, 
\emph{$L_p$-universality} can be related to all other notions considered in this paper, through Figure~\ref{Fig:results}. 

\appendix
\newcommand{\appsection}[1]{\let\oldthesection\thesection
  \renewcommand{\thesection}{Appendix \oldthesection}
  \section{#1}\let\thesection\oldthesection}
\acks{B. K. S. and G. R. G. L. wish to acknowledge support from the Institute of Statistical Mathematics (ISM), Tokyo, the National Science Foundation (grant DMS-MSPA 0625409), the Fair Isaac Corporation and the University of California MICRO program. Part of this work was done while B. K. S. was visiting ISM. K.F. was supported by JSPS KAKENHI 19500249.}

\appendix
\section*{Appendix A. Supplementary Results}\label{appendix-a}
For completeness, we present the following supplementary result, which 
is a simple generalization of the technique used in the proof of \citet[Theorem 3]{Sriperumbudur-08}.
\begin{lemma}\label{lem:interchange}
Let $k$ be a measurable and bounded kernel on a measurable space, $X$ and let $\eu{H}$ be its associated RKHS. Then, for any $f\in\eu{H}$ and for any finite signed Borel measure, $\mu$, 
\begin{equation}\label{Eq:interchange}
\int_X f(x)\,d\mu(x)=\int_X \langle f,k(\cdot,x)\rangle_\eu{H}\,d\mu(x)=\Big\langle f,\int_X k(\cdot,x)\,d\mu(x)\Big\rangle_\eu{H}.
\end{equation}
\end{lemma}
\begin{proof}
Let $T_\mu:\eu{H}\rightarrow\bb{R}$ be a linear functional defined as $T_\mu[f]:=\int_X f(x)\,d\mu(x)$. It is easy to show that 
\begin{equation}
\Vert T_\mu\Vert:=\sup_{f\in\eu{H}}\frac{|T_\mu[f]|}{\Vert f\Vert_\eu{H}}\le \sqrt{\sup_{x\in X}k(x,x)}\Vert\mu\Vert<\infty.\nonumber
\end{equation}
Therefore, $T_\mu$ is a bounded linear functional on $\eu{H}$. By the Riesz representation theorem \cite[Theorem 5.25]{Folland-99}, there exists a unique $\lambda_\mu\in\eu{H}$ such that $T_\mu[f]=\langle f,\lambda_\mu\rangle_\eu{H}$ for all $f\in\eu{H}$. Set $f=k(\cdot,u)$ for some $u\in X$, which implies $\lambda_\mu=\int_X k(\cdot,x)\,d\mu(x)$ and the result follows.\vspace{-2mm}
\end{proof}

\vskip 0.2in

\end{document}